\pdfoutput=1

\documentclass[11pt]{article}

\usepackage{acl}

\usepackage{times}
\usepackage{latexsym}

\usepackage[T1]{fontenc}

\usepackage[utf8]{inputenc}

\usepackage{microtype}
\usepackage{booktabs}

\usepackage{graphicx}
\usepackage{enumitem}
\usepackage{caption}
\usepackage{subcaption}
\usepackage{floatrow}
\usepackage{placeins}

\newcommand{\customfootnotetext}[2]{{
  \renewcommand{\thefootnote}{#1}
  \footnotetext[0]{#2}}}

\newcommand{\anas}[1]{{#1}}

\title{Exploring The Landscape of Distributional Robustness \\ for Question Answering Models}

\author{Anas Awadalla$^1$
        \And
  Mitchell Wortsman$^1$
        \And
  Gabriel Ilharco$^1$\hspace*{-1.5em}
        \And
  Sewon Min$^1$
        \AND
  Ian Magnusson$^2$
        \And 
  Hannaneh Hajishirzi$^{1,2}$
        \And
  Ludwig Schmidt$^{1,2}$
  }

\begin{document}
\maketitle
\customfootnotetext{1}{University of Washington. $^2$Allen Institute for AI.}
\customfootnotetext{ }{Correspondence to 
\texttt{anasa2@cs.washington.edu}.}

\begin{abstract}
We conduct a large empirical evaluation to investigate the landscape of distributional robustness in question answering. Our investigation spans over 350 models and 16 question answering datasets, including a diverse set of architectures, model sizes, and adaptation methods (e.g., fine-tuning, adapter tuning, in-context learning, etc.). We find that, in many cases, model variations do not affect robustness and in-distribution performance alone determines out-of-distribution performance.
Moreover, our findings indicate that
i) zero-shot and in-context learning methods are more robust to distribution shifts than fully fine-tuned models;
ii) few-shot prompt fine-tuned models exhibit better robustness than few-shot fine-tuned span prediction models;
iii) parameter-efficient and robustness enhancing training methods provide no significant robustness improvements.
In addition, we publicly release all evaluations to encourage researchers to further analyze robustness trends for question answering models.

\end{abstract}

\section{Introduction}
\begin{figure*}
\floatbox[{\capbeside\thisfloatsetup{capbesideposition={right},capbesidewidth=6cm}}]{figure}[\FBwidth]
{\caption{
    We evaluate over 350 models on 16 datasets to characterize the landscape of distributional robustness in question answering.
    Our results span a variety of architectures and adaptation strategies, including zero-shot inference, fine-tuning, and in-context learning (ICL).
    The $x$-axis shows performance on SQuAD (in-distribution), while the $y$-axis shows the average performance on the 15 other QA datasets (out-of-distribution).
    Almost all models lie under the $y = x$ diagonal, i.e., performance drops under distribution shift.
    Moreover, within certain groups of models---for instance, ICL models---in-distribution performance accurately predicts out-of-distribution performance.
    As in~\citet{taori2020measuring}, we apply logit axis scaling to clarify that the relationship between in-distribution and out-of-distribution performance is approximately linear in the logit domain.
    }\label{fig:data}}
{\includegraphics[width=9cm]{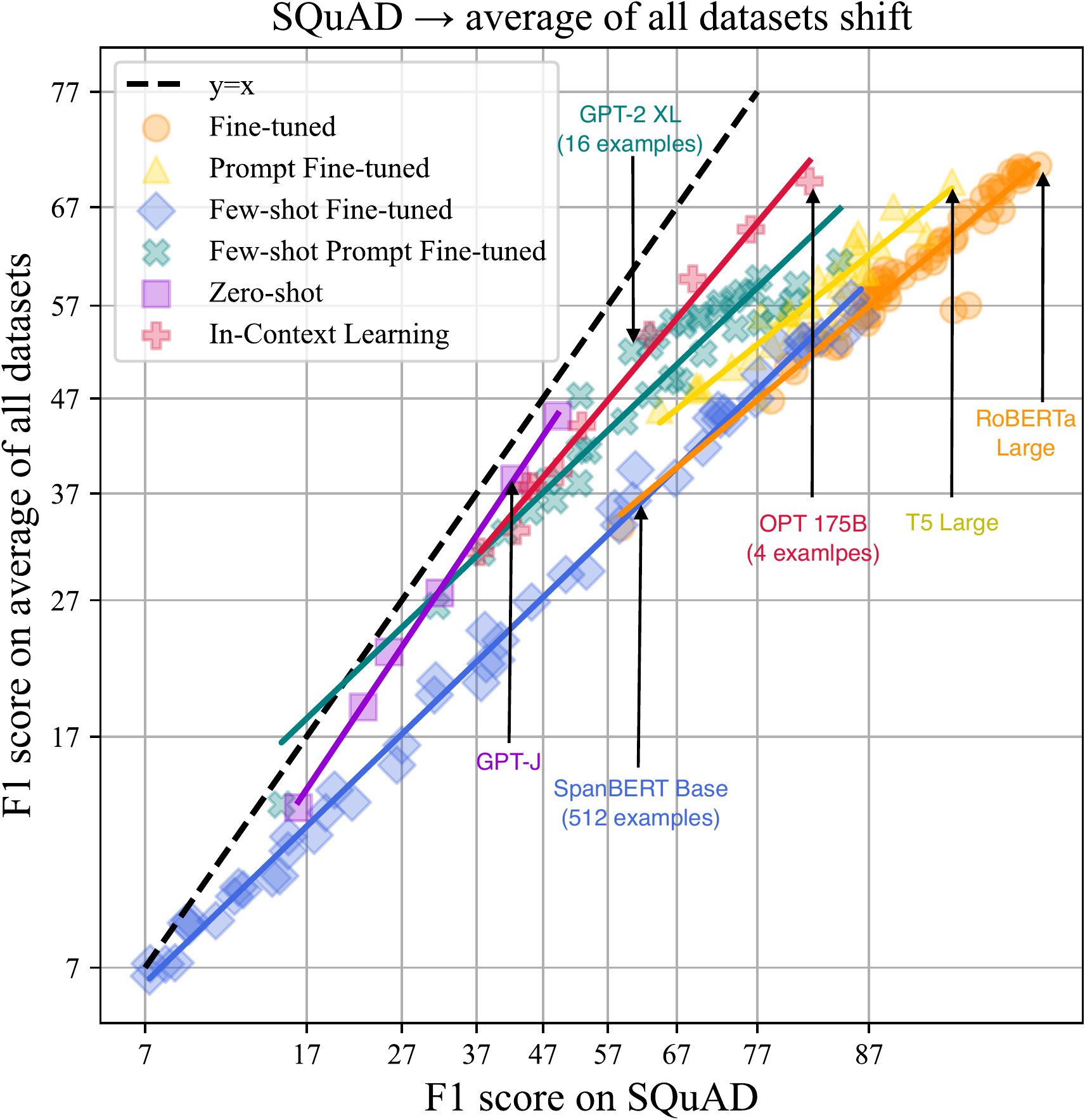}}
\vspace*{-0.3cm}
\end{figure*}
Over the past few years, natural language processing has seen substantial progress.
In many benchmarks, large pre-trained models adapted to a target dataset reach or even surpass human performance \cite[inter alia]{devlin-etal-2019-bert,raffel2019exploring,Radford2019LanguageMA,brown2020language,hoffmann2022training,chowdhery2022palm}.
At the same time, current methods still fail to generalize reliably in a variety of test conditions \cite{ribeiro-etal-2020-beyond,gardner-etal-2020-evaluating,koh2021wilds,luu2021time,ribeiro-lundberg-2022-adaptive}, which limits their applicability and raises questions about what exactly the methods learn \cite{bender-koller-2020-climbing}. 
One limitation of current benchmarks is that they often measure performance only on data that comes from the same distribution as the training set \cite{wang-etal-2018-glue,wang2019superglue}.
However, evaluating models on a single test set provides no information on whether a method also performs well under distribution shift.
While there is an increasing amount of research on robustness in NLP \cite[inter alia]{ribeiro-etal-2020-beyond,tu-etal-2020-empirical,hendrycks-etal-2020-pretrained,gardner-etal-2020-evaluating,arora-etal-2021-types,veitch2021counterfactual,goel2021robustness,Miller2020TheEO}, the community has not yet adopted a common set of best practices for evaluating robustness.
As a result, new methods often do not evaluate on comparable or even any robustness test sets, which makes it challenging to understand which methods generalize more reliably and whether NLP is making progress on robustness to distribution shift.

To address this challenge and shed light on the robustness landscape in NLP, we conduct a large empirical evaluation of distributional robustness in question answering (QA).
Building on recent research on robustness in computer vision \cite{taori2020measuring,miller2021accuracy}, we focus on distribution shifts that arise between two related but different test sets.
These distribution shifts are sometimes called dataset shift to distinguish them from other kinds of distribution shift.
An example of dataset shift is a pair of QA test sets where one test set is constructed from Wikipedia articles and the other from Amazon product reviews, possibly also with a different crowdsourcing process.
In contrast to other notions of robustness such as adversarial robustness, dataset shifts involve no synthetic perturbations of existing test examples and are therefore more representative of generalization challenges arising ``in the wild'' \cite{taori2020measuring}.

Within the scope of dataset shifts for QA, our robustness evaluation includes a wide range of models and distribution shifts.
Specifically, we assembled a testbed of over 350 QA models and 16 QA datasets, including SQuAD v1.1~\cite{rajpurkar-etal-2016-squad}, SquadShifts~\cite{Miller2020TheEO}, and MRQA test sets~\cite{fisch2019mrqa}.
Our testbed spans different model architectures, model sizes, and pre-training setups.
In addition, we evaluate a variety of approaches for applying pre-trained models to question answering including supervised fine-tuning, in-context learning, parameter-efficient fine-tuning, zero-shot inference, and more.
Finally, we also include methods specifically designed to enhance robustness such as RXF~\cite{aghajanyan2021better} and FreeLB~\cite{Zhu2020FreeLB}.

Our testbed enables us to both identify overarching trends spanning many models, and to contextualize the robustness behavior of individual models.
Among our findings are the following key results:
\begin{itemize}
\itemsep.2em 
\item Dataset shift still is an unsolved problem in QA: most models suffer a large performance drop under this kind of distribution shift.

\item Despite different architectures and model sizes, many models follow a consistent trend relating in-distribution and out-of-distribution performance. Improving in-distribution performance usually also increases out-of-distribution performance in a predictable way.

\item Current robustness interventions follow the same trend as models without such interventions, i.e., the robustness interventions do not increase robustness to dataset shifts.

\item The only exception to the otherwise universal performance trend are zero-shot, in-context learning, and few-shot prompt fine-tuned models. These models are more robust than the baseline given by the other models in our testbed. However, the robustness of large decoder-only models decreases as the models are fine-tuned on more data from the target task.
\end{itemize}

Figure~\ref{fig:data} summarizes our findings and shows the average F1 score on all distribution shifts as a function of the F1 score on SQuAD.
Interestingly, our overall results are analogous to similar large-scale robustness evaluations in computer vision~\cite{taori2020measuring,miller2021accuracy,Radford2021LearningTV}, which suggests that there may be a shared underlying mechanism behind these distribution shifts that warrants further investigation.

We hope that our work helps clarify the state of robustness in NLP and provides a starting point for future work.
To simplify measuring robustness to dataset shift and enable future robustness improvements, we will release our testbed including all 350+ models and evaluation results.

The remainder of the paper is organized as follows: first, we detail background and experimental setup (\S\ref{sec:experimental_setup}). Next, we introduce and answer our specific research questions (\S\ref{sec:results}, \ref{sec:discussion}).
Finally, we discuss the limitations of our approach, overall conclusions, and directions for future investigation (\S\ref{sec:conclusion}, \ref{sec:limitations}).

\begin{figure}
    \centering
    \vspace*{-2cm}
    \includegraphics[width=\columnwidth]{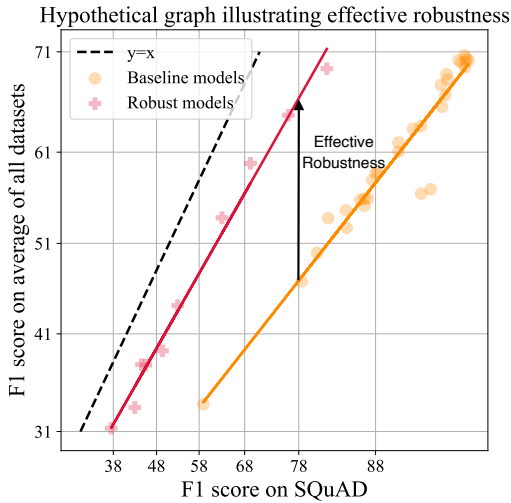}
    \vspace*{-2cm}
    \caption{A schematic which illustrates the robustness measuring technique we use. \emph{Effective robustness scatter plots}~\cite{pmlr-v97-recht19a, taori2020measuring} display performance on the distribution from which training data is from (in-distribution) on the $x$-axis, and out-of-distribution performance on the $y$-axis.
    Effective robustness is vertical movement towards the $y=x$ diagonal beyond the baseline trend fit to fully fine-tuned models---a model with higher effective robustness has more consistent performance in- and out-of-distribution.}
    \label{fig:example}
    \vspace*{-0.2cm}
\end{figure}

\section{Experimental Setup}
\label{sec:experimental_setup}

Our testbed includes over 350 models, covering a broad range of model architectures, pre-training datasets, and adaptation strategies.
We use SQuAD v1.1~\cite{rajpurkar-etal-2016-squad} as our reference point for question answering performance because SQuAD is a popular dataset and the performance ceiling is comparatively well understood since humans can achieve an F1 score around 95~\citep{Miller2020TheEO}.
For all models except those performing zero-shot inference, we adapt the models to question answering with the SQuAD training set.

We evaluate robustness to distribution shift on the remaining 15 question answering datasets (Table~\ref{dataset info}).
We follow \citet{taori2020measuring} in defining robustness, i.e., we say a model is robust if it has consistent performance under a distribution shift from a reference distribution to another distribution.
We refer to SQuAD as in-distribution (ID) and the other 15 datasets as out-of-distribution (OOD).
In the remainder of this section, we describe the different models, adaptation strategies, datasets, and evaluation details.

\begin{table*}[h]
\setlength{\tabcolsep}{4pt}
\centering
\footnotesize
\begin{tabular}{ lrl }
\toprule
Dataset name &Test set size & Domains \\
\midrule
SQuAD v1.1 dev. set \cite{rajpurkar-etal-2016-squad}& 10,570    & Wikipedia\\
 SquadShifts New-Wiki \cite{Miller2020TheEO}&7,938  & Wikipedia\\
 SquadShifts Reddit \cite{Miller2020TheEO}&9,803 & Reddit\\
 SquadShifts NYT \cite{Miller2020TheEO}&10,065 & New York Times\\
 SquadShifts Amazon \cite{Miller2020TheEO}&   9,885  & Amazon reviews\\
 RACE \cite{lai-etal-2017-race}& 674  & English exams from China \\
 DROP \cite{dua-etal-2019-drop}& 1,503  & Wikipedia\\
 NewsQA \cite{trischler-etal-2017-newsqa}& 4,212  & CNN articles\\
 SearchQA \cite{Dunn2017SearchQAAN}& 16,980  & Jeopardy! questions with contexts from Google search \\
 NaturalQuestions \cite{kwiatkowski-etal-2019-natural}& 12,836  & Google search questions with contexts from Wikipedia\\
 DuoRC (ParaphraseRC) \cite{Saha2018DuoRCTC}& 1,501  & Movie plots from IMDB and Wikipedia\\
 HotpotQA \cite{yang-etal-2018-hotpotqa}& 5,904  & Wikipedia\\
 TextbookQA \cite{Kembhavi_2017_CVPR}& 1,503  & Middle school science questions from textbooks\\
 TriviaQA \cite{joshi-etal-2017-triviaqa}& 7,785  & Trivia questions with contexts collected using a Bing search\\
 RelationExtraction \cite{levy-etal-2017-zero}& 2,948  & Generated samples using a knowledge base \\
 BioASQ \cite{TsatsaronisBioasq}& 1,504  &  Medical articles\\
\bottomrule
\end{tabular}
\caption{Question answering datasets used to evaluate models in this work.
SQuAD is used as the in-distribution reference dataset---we use training data from SQuAD to adapt models.
The remaining datasets are used to answer the question of how SQuAD models perform under dataset shift---we use these other datasets for evaluation only.}
\label{dataset info}
\vspace*{-0.3cm}
\end{table*}

\subsection{Models}
Our testbed focuses on transformer models ranging from 11 million to 175 billion parameters.
We explore several encoder-only models---ALBERT~\cite{Lan2020ALBERT}, BERT~\cite{devlin-etal-2019-bert}, SpanBERT~\cite{joshi-etal-2020-spanbert}, RoBERTa~\cite{liu2019roberta}, and Splinter~\cite{Ram2021FewShotQA}---encoder-decoder models ---T5~\cite{raffel2019exploring} and BART~\cite{lewis-etal-2020-bart}---and decoder-only models (GPT-2~\cite{Radford2019LanguageMA}, OPT~\cite{zhang2022opt}, GPT-Neo~\cite{gpt-neo}, and GPT-J~\cite{gpt-j}).

\subsection{Adaptation strategies}
We evaluate multiple adaptation strategies---methods that {\em adapt} the pre-trained language model to perform better on a downstream task using labeled, in-distribution training data, e.g., through gradient based learning and in-context learning.
We also examine models evaluated in a zero-shot setting, which we also refer to as an adaption method for consistency, even though no data from the in-distribution dataset is observed.
For a subset of these models we also explore few-shot instead of full-shot adaptation to assess the impact of the number of training examples on robustness. 

\subsubsection{\anas{Fine-tuning (baseline)}}
We include a common fine-tuning method: \anas{adding a span prediction head and} updating all the parameters in a language model via additional training on a downstream dataset, as done in \citet{devlin-etal-2019-bert} and subsequent work.

\anas{\subsubsection{Prompt fine-tuning}
Prompt fine-tuning adds no additional task specific layers and fine-tunes the existing weights to generate the answer. We use next token prediction when fine-tuning auto-regressive models like GPT. For T5 and BART models we use two fine-tuning tasks: 1) casting QA as an infilling task and generate the answer by predicting a masked span 2) conditioning the model on the context and question and fine-tune it to generate the answer.
}
\subsubsection{Parameter-efficient fine-tuning}
Parameter-efficient fine-tuning modifies only a small percentage of existing or auxiliary parameters, while freezing all other parameters. We evaluate Houlsby~\cite{Houlsby2019ParameterEfficientTL} and Pfeiffer~\cite{pfeiffer-etal-2021-adapterfusion} adapters, prefix tuning~\cite{li-liang-2021-prefix}, and LoRA~\cite{hu2021lora}. 
While these methods modify only a small number of parameters, they have been shown to be competitive with full fine-tuning when measuring in-distribution performance.
Previous work suggests freezing a majority of model weights may make these methods more robust~\cite{lester-etal-2021-power}.



\subsubsection{Robustness enhancing fine-tuning}
We evaluate methods which have been designed to improve model robustness.
In particular, we evaluate  RXF~\cite{aghajanyan2021better} and FreeLB~\cite{Zhu2020FreeLB}, which apply adversarial training strategies to improve generalization.
Previous work evaluated robustness by comparing only to a few models and do not run extensive evaluations in question answering. Our work conducts evaluations on a large number of distribution shifts.

\vspace*{-0.1cm}
\subsubsection{In-context learning}
In-context learning is an adaptation method proposed by \citet{NEURIPS2020_1457c0d6} that does not require any gradient updates. This is particularly useful for very large language models, where fine-tuning is expensive.
In-context learning refers to the process of conditioning a language model on one or more samples from a training set at inference time, allowing the model to perform a task without updating any parameters. For our experiments, we condition the model on triplets of context, question, and answer, as in  \citet{NEURIPS2020_1457c0d6}. 



\vspace*{-0.1cm}
\subsubsection{Zero-shot inference}
We evaluate models using prompting or zero-shot inference~\cite{Radford2019LanguageMA}, where a model is conditioned only on the context and question of each test example. In other words, the model generates an answer without conditioning on training examples. 
Zero-shot models do not observe data from the reference distribution and have been shown to exhibit consistent performance across many distributions in computer vision~\cite{Radford2021LearningTV}.




\subsection{Distribution shifts}

We consider models which are trained on a reference distribution, which we also refer to as the in-distribution, with the exception of zero-shot models.
In addition to measuring model performance on this reference distribution, we also evaluate model performance on other datasets where data distribution changes from the reference distribution.
We refer to these other datasets as out-of-distribution, and we are interested in model behavior under distribution shift.
Concretely, we want to measure how model performance changes when evaluated in- and out-of-distribution.

While there is extensive literature studying adversarial distribution shifts~\cite{wu-etal-2021-evaluating}, our work focuses on \textit{natural} distribution shifts \cite{taori2020measuring}, where the out-of-distribution datasets are not generated via synthetic perturbations to existing datasets.

In this work, we use the popular SQuAD~\cite{rajpurkar-etal-2016-squad} dataset as the reference (in-distribution) dataset.
In addition, we evaluate model performance on 15 out-of-distribution datasets.
We choose SQuAD as the reference distribution as it is one of the largest and the most well-studied QA datasets.

For our out-of-distribution test sets, we use the four datasets presented in the SquadShifts~\cite{Miller2020TheEO} in addition to datasets from the MRQA~\cite{fisch2019mrqa} testbed. Details about each of these datasets can be found in Table~\ref{dataset info}.


\subsection{Measuring robustness}
We follow the technique for measuring model robustness that is outlined in~\citet{taori2020measuring}: a model is said to be robust if it exhibits consistent performance in- and out-of-distribution.
This is advantageous compared to examining only out-of-distribution performance because it removes the 
confounder of in-distribution performance (as shown in \cite{taori2020measuring, pmlr-v139-miller21b}, models which achieve better performance in-distribution will often also perform better out-of-distribution).

As in \citet{taori2020measuring}, the robustness measure we consider can be illustrated by looking at a scatter plot.
For an illustrated example of this we refer to Figure~\ref{fig:example}, which displays the F1 score on the SQuAD development set on the $x$-axis and the F1 score averaged over the out-of-distribution datasets on the $y$-axis.
Each point on the scatter plot is a different model.
\emph{Effective robustness} then describes \emph{vertical} movement in this scatter plot towards the $y=x$ line.
In particular, effective robustness measures performance out-of-distribution beyond the trend fit to fully fine-tuned models.
This vertical movement is movement towards a model that has consistent performance in- and out-of-distribution (i.e., on aggregate fully fine-tuned models have $\sim$0 effective robustness).
In Figure~\ref{fig:example}, which schematizes results that we will later observe with real data, models that are more robust sit above the baseline trend and exhibit robustness---the models shown in orange are more robust than the other models as they have better out-of-distribution performance given the same in-distribution performance. 
\begin{figure}
    \centering
    \includegraphics[width=\columnwidth]{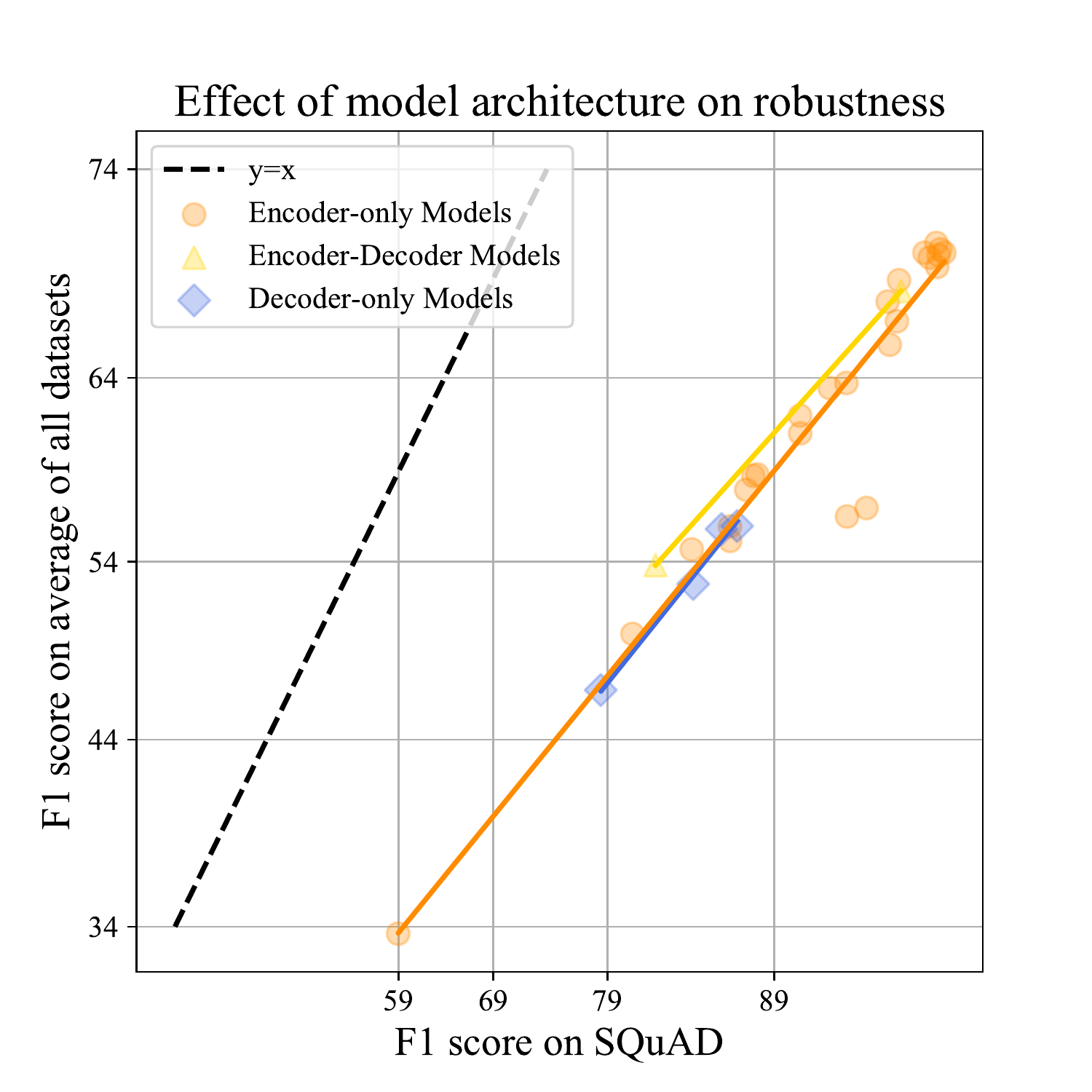}
    \caption{
    \anas{Encoder-only, encoder-decoder, and decoder-only models are equally as robust when fine-tuned by adding a span prediction head. We conclude that architecture does not determine distributional robustness.}
    }
    \label{fig:arch}
\end{figure}

\begin{figure}
    \centering
    \includegraphics[width=\columnwidth]{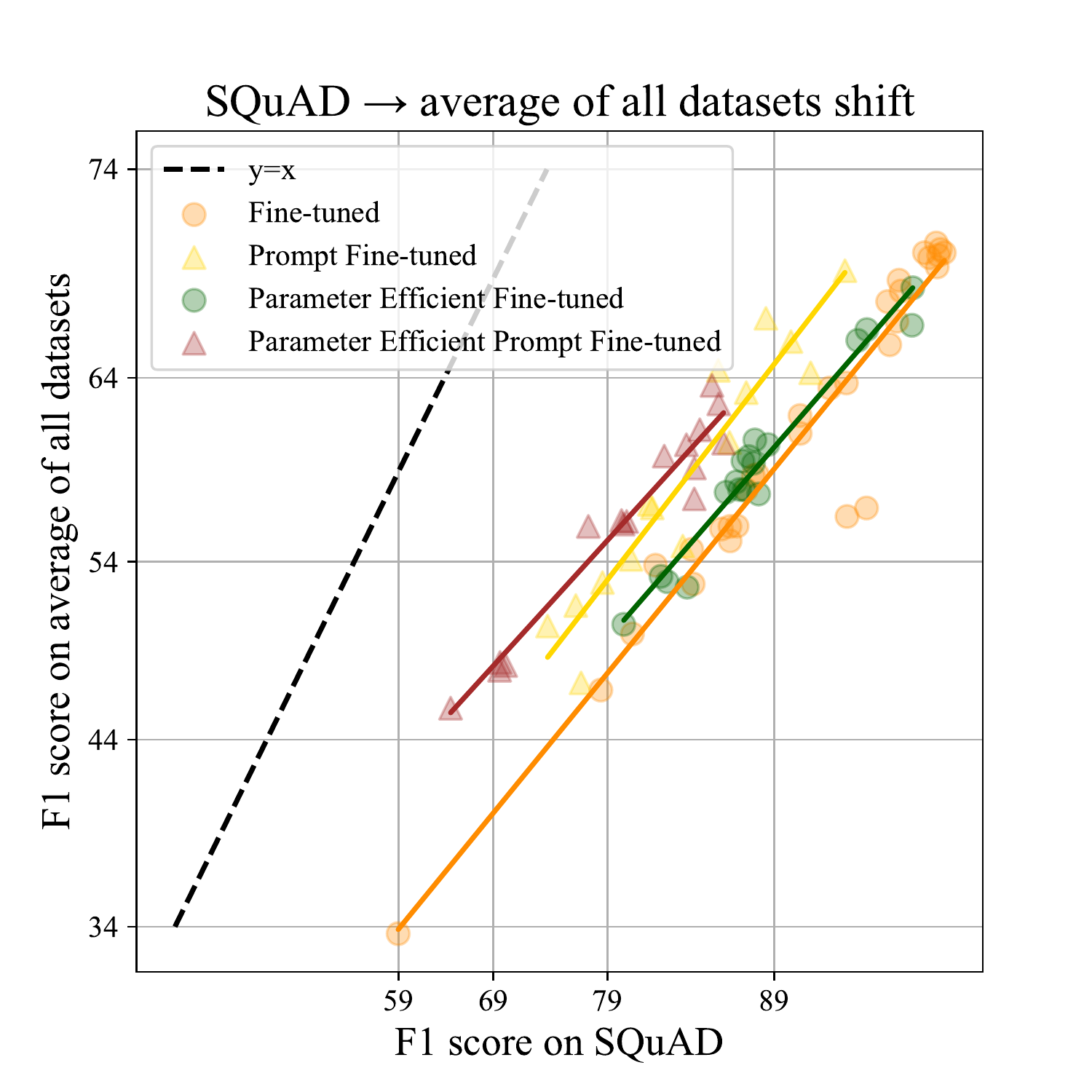}
    \caption{Parameter-efficient fine-tuning methods (highlighted in red and green) do not exhibit noticeable robustness improvements compared to other fine-tuned models.
    }
    \label{fig:param-eff}
\end{figure}
\section{Results} 
\label{sec:results}
This section aims to answer our main research questions: 
\begin{itemize}[wide, labelwidth=0pt, labelindent=0pt]
    \item How do models perform under distribution shift? 
    \item Are some models more robust than others? 
    \item Do adaptation methods impact robustness? 
\end{itemize}
We answer these questions in Sections \ref{sec:30}, \ref{sec:31} and \ref{sec:32}, respectively.

\subsection{Performance drops under distribution shift}\label{sec:30}

As shown in Figure~\ref{fig:data}, we observe that model performance drops under distribution shift.
This effect is more pronounced for the best models on SQuAD, which are fully fine-tuned.
This indicates that, despite progress in question answering, there is still substantial room for progress in improving model robustness.

\subsection{Role of model}\label{sec:31}
\paragraph{Role of model architecture.}
In Figure~\ref{fig:arch} we compare the robustness of fine-tuned encoder-only, decoder-only, and encoder-decoder architectures. \anas{Our experiments indicate that architecture does not impact robustness. We observe that when different model families are adapted using a span prediction head, all models are equally robust.}
One limitation in our comparison is that the architectures we compare do not share the same pre-training corpus. However, larger corpora have been shown to improve robustness in computer vision~\cite{Radford2021LearningTV}. This is an area that could be investigated further in future work.

\begin{figure}
    \centering
    \includegraphics[width=\columnwidth]{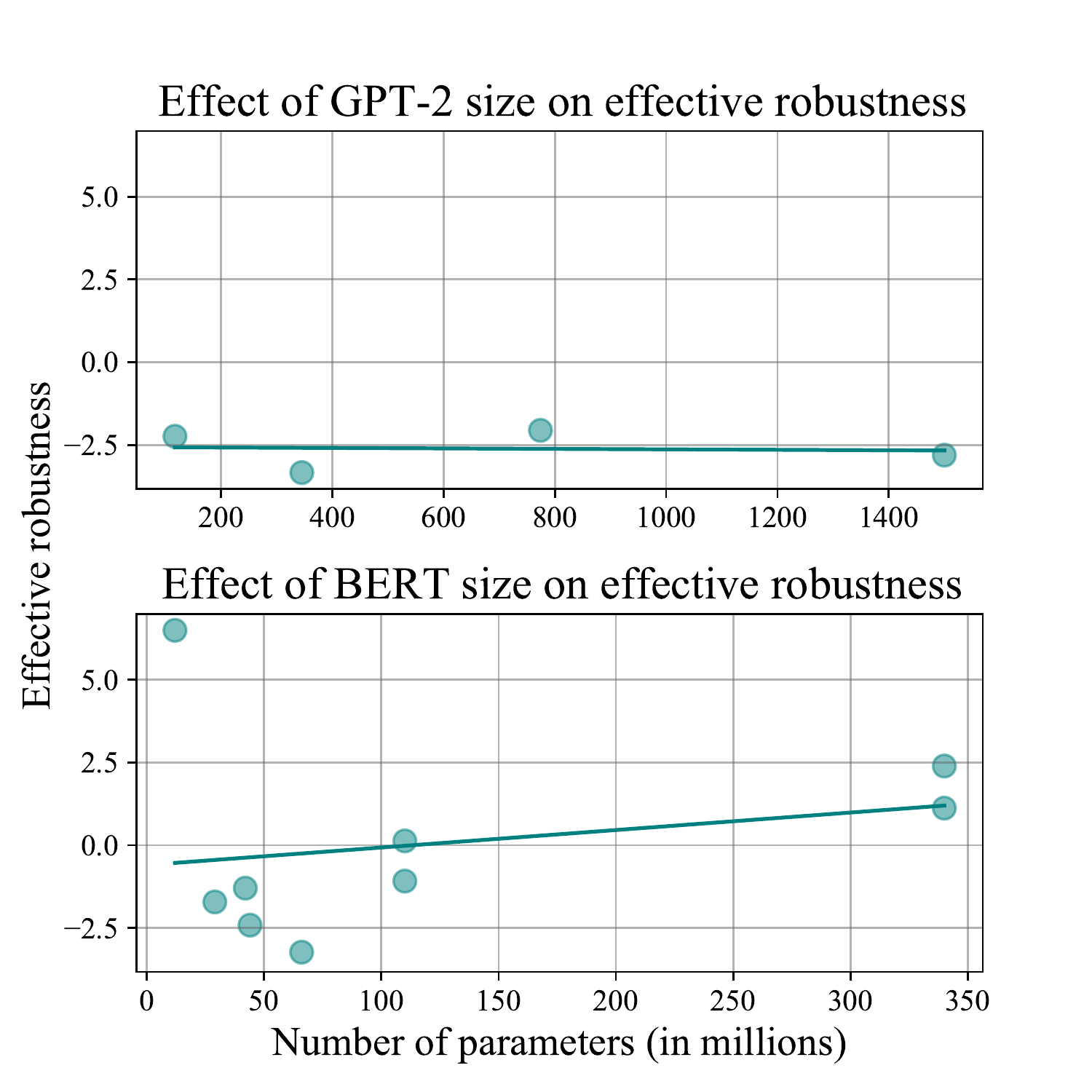}
    \caption{\anas{Average effective robustness of BERT and GPT-2 as a function of the number of parameters of models fine-tuned on SQuAD.
    Overall model size does not determine robustness.
    }}
    \label{fig:model size}
\end{figure}

\paragraph{Role of model size.} Previous work~\cite{hendrycks-etal-2020-pretrained} has claimed that model size does not affect the robustness of language models. In Figure~\ref{fig:model size} we plot the average \emph{effective robustness} on all distribution shifts as a function of the number of model parameters for \anas{fine-tuned GPT-2 and BERT models to control for pre-training corpus and architecture. Overall, we observe that model size is not strongly correlated with robustness.} 

\subsection{Role of the adaptation method} \label{sec:32}
\begin{figure}
    \centering
    \includegraphics[width=\columnwidth]{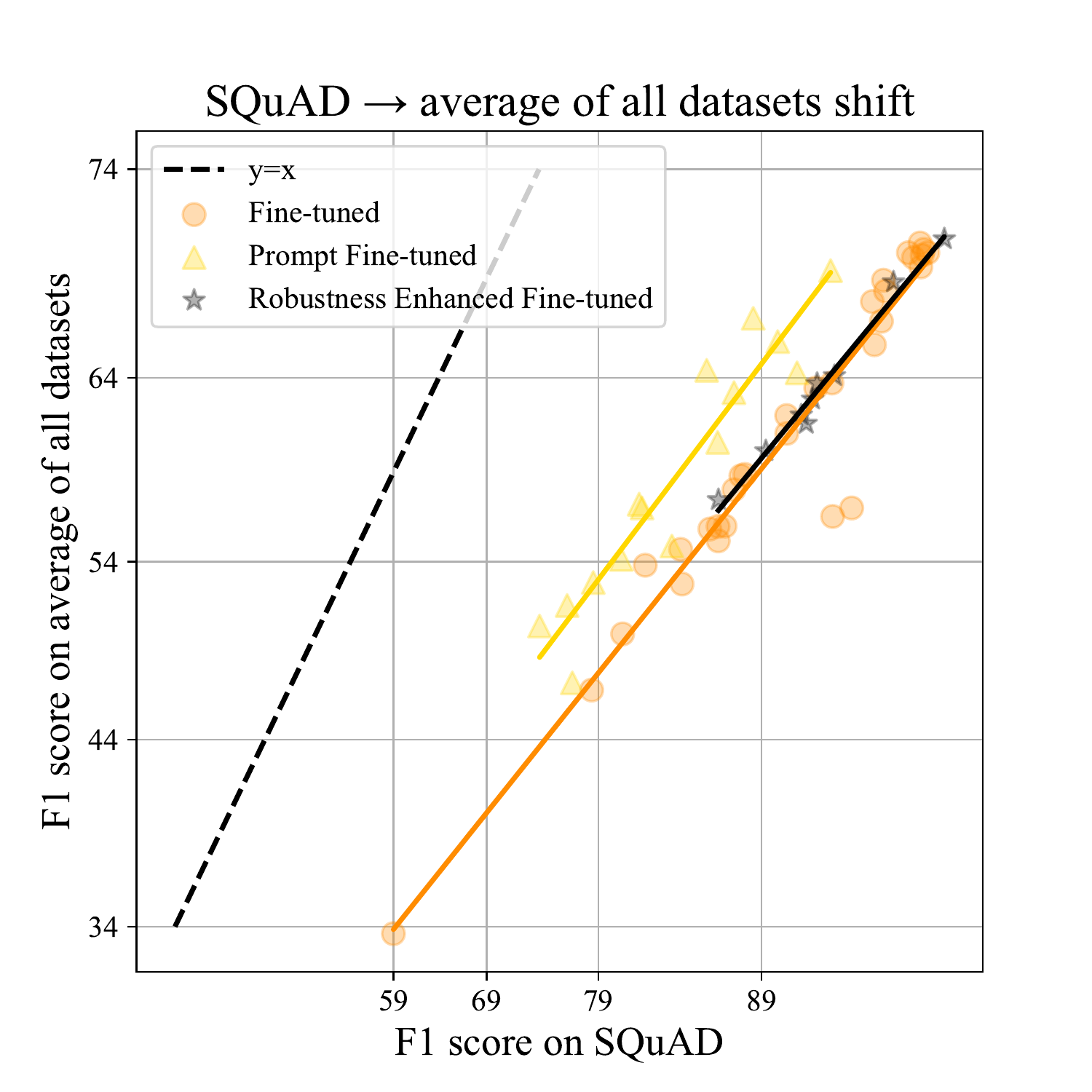}
    \caption{
    Methods designed to improve robustness (highlighted in black) do not exhibit noticeable robustness improvements on our testbed.
    This discrepancy may arise because of our focus on question answering, which previous work does not evaluate on.
    }
    \label{fig:robustness}
\end{figure}

\paragraph{Zero-shot and in-context learning (ICL).}
We find that both zero-shot and in-context learning methods exhibit more robustness than methods that use gradient-based learning.
As illustrated by Figure~\ref{fig:data}, the trend for zero-shot and in-context learning models is well above the trend of all other models.
This entails that for the same in-distribution performance, we expect better out-of-distribution performance for in-context learning and zero-shot inference
\begin{figure}
    \centering
    \includegraphics[width=\columnwidth]{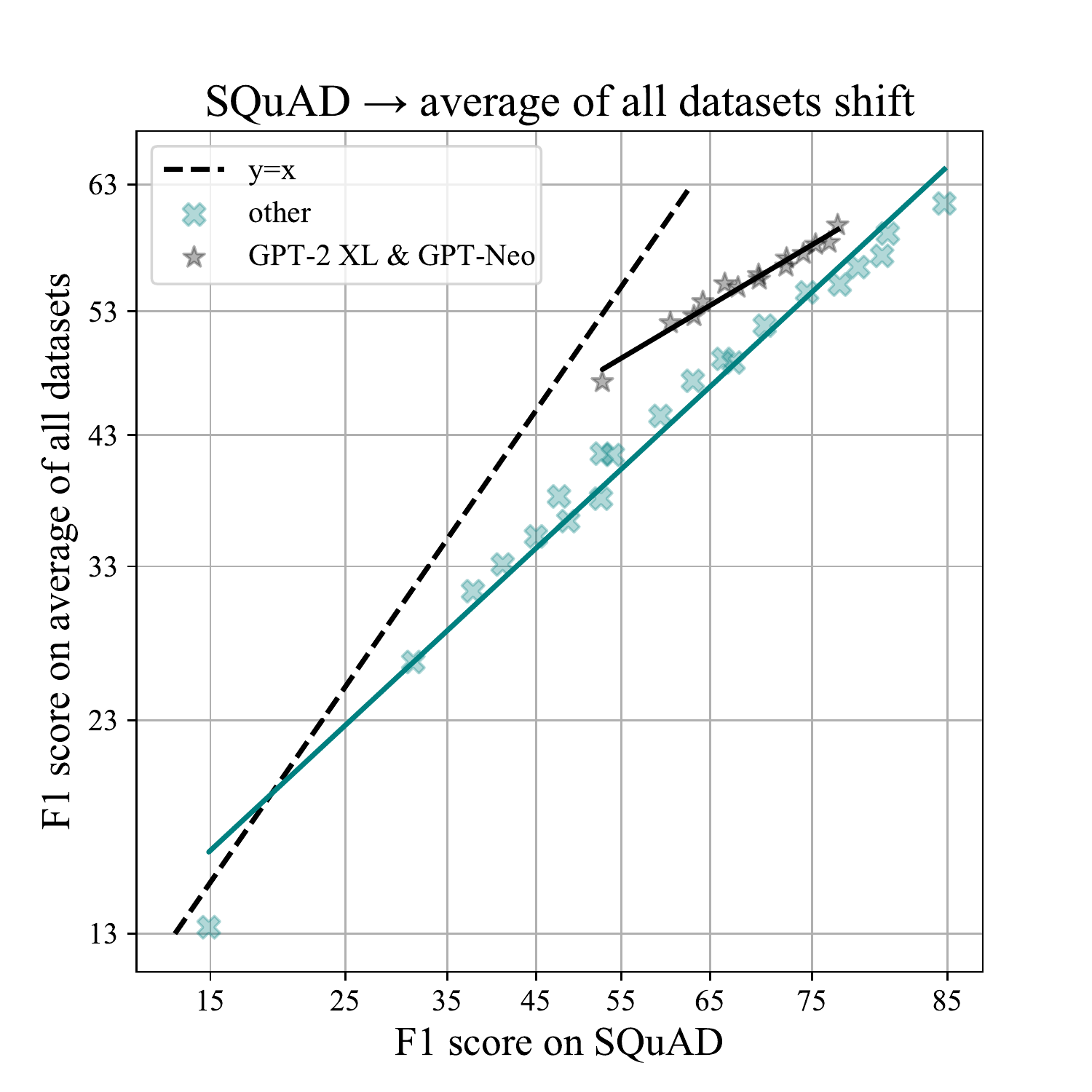}
    \caption{Few-shot prompt fine-tuned billion parameter GPT models (colored black) are more robust than smaller few-shot prompt fine-tuned models. Further investigation is required to determine if the increase in effective robustness is due to architecture or model size.}
    \label{fig:gpt robustness}
\end{figure}
\paragraph{Few-shot fine-tuning.} In Figure~\ref{fig:data}, we observe that 
\anas{few-shot methods follow two separate robustness trends.}
\anas{\begin{enumerate}
    \item Few-shot fine-tuned models are on a trend similar to fully fine-tuned models.
    \item Few-shot prompt fine-tuned models are more robust than all other models that use gradient based learning.
\end{enumerate}}

\begin{figure}
    \centering
    \includegraphics[width=\columnwidth]{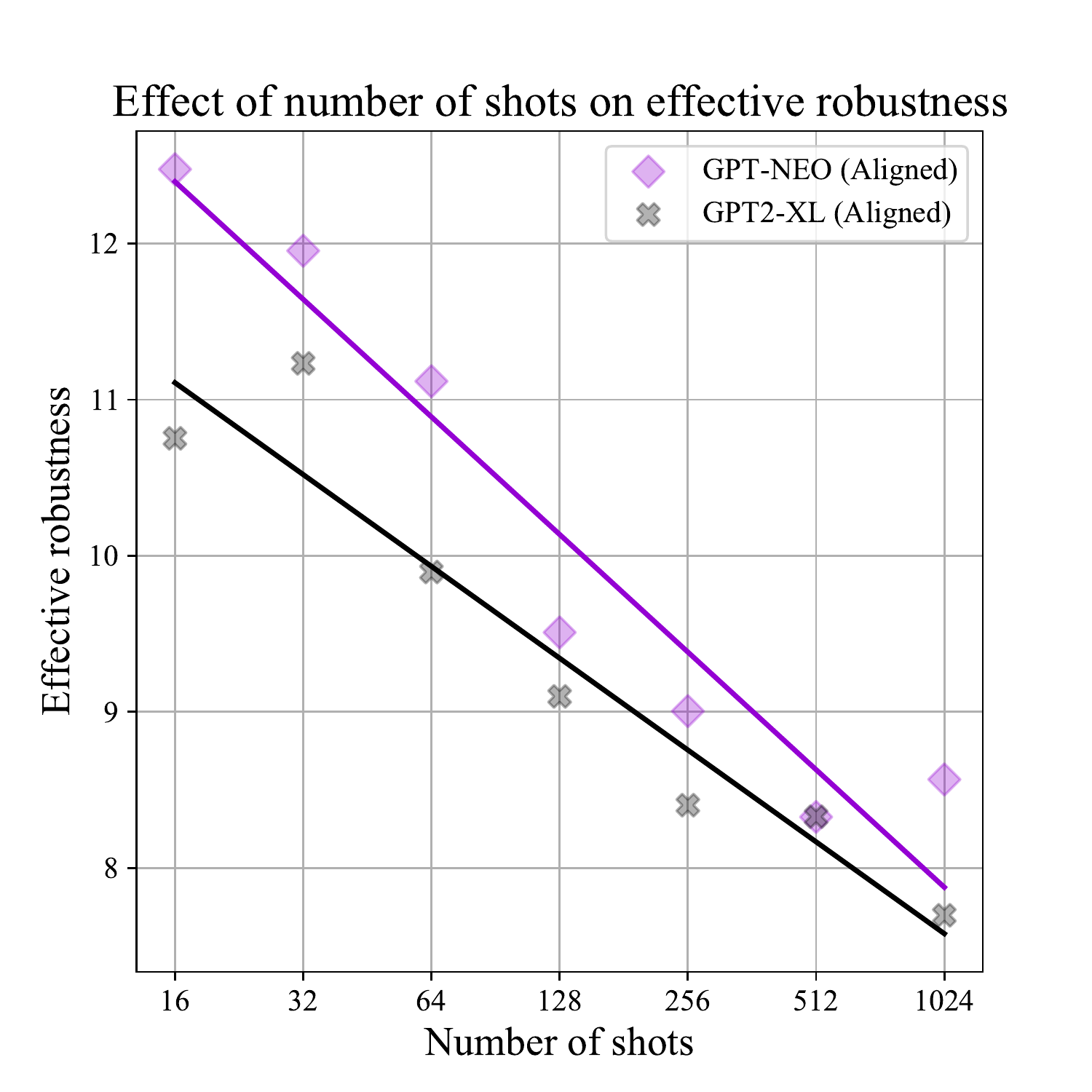}
    \caption{Average effective robustness for each GPT model as a function of the number of shots used for fine-tuning. As the number of shots increases the model becomes better in distribution but the average effective robustness decreases.}
    \label{fig:effective robustness}
    \vspace*{-0.2cm}
\end{figure}

\anas{Notable outliers to the few-shot prompt fine-tuned model trend are the GPT-2 XL~\cite{Radford2019LanguageMA} and GPT-Neo 1.3B~\cite{gpt-neo} models. As shown in Figure~\ref{fig:gpt robustness}, these models are more robust than other few-shot prompt fine-tuned models.} This indicates that models with better zero-shot capabilities can generalize better when fine-tuned in the few-shot setting.
For these few-shot fine-tuned GPT models we explore how the number of training shots impacts robustness.
We find that as the number of training samples increases, the effective robustness of few-shot GPT models decreases as shown in Figure~\ref{fig:effective robustness}. In particular, increasing the number of shots from 16 to 1024 decreases effective robustness.
This observation interpolates our previous results: a GPT model used in the zero-shot setting is robust while prompt fine-tuned GPT models are less robust.
As observed by previous work~\cite{Radford2021LearningTV,andreassen2021evolution,wortsman2021robust}, fine-tuning a model can reduce robustness and lead to a model which is overspecialized to the downstream task.
\paragraph{Full fine-tuning using span prediction.} The fully fine-tuned models exhibit noticeably less robustness than other adaptation methods, however they also have the best performance on SQuAD. The best performing model on SQuAD has similar performance out-of-distribution to the best ICL model, despite performing more than 10 percentage points better in-distribution.
\paragraph{Fine-tuning using a prompt.} We find that prompt fine-tuning methods are more robust in comparison to fine-tuned models. We observe that not using span prediction and instead fine-tuning existing model weights to generate the answer allows the model to maintain some of robustness from the zero-shot setting.


\paragraph{Parameter-efficient tuning.}
We examine the performance of parameter-efficient fine-tuning methods for
different architectures and model sizes.
Our results indicate that these methods are neither noticeably more robust or less robust than fine-tuning all parameters \anas{when using prompt based methods or span prediction, as shown in} 
Figure~\ref{fig:param-eff}.
\paragraph{Methods designed to enhance robustness.}
As illustrated by Figure~\ref{fig:robustness} we find that RXF and FreeLB, which are designed to improve robustness,
do not exhibit noticeable robustness improvements on the distribution shifts.
We believe that one of the values of our large test bed is to comprehensively evaluate future robustness enhancing methods.

\begin{figure*}
     \centering
     \begin{subfigure}[b]{0.32\textwidth}
         \centering
         \includegraphics[width=\textwidth]{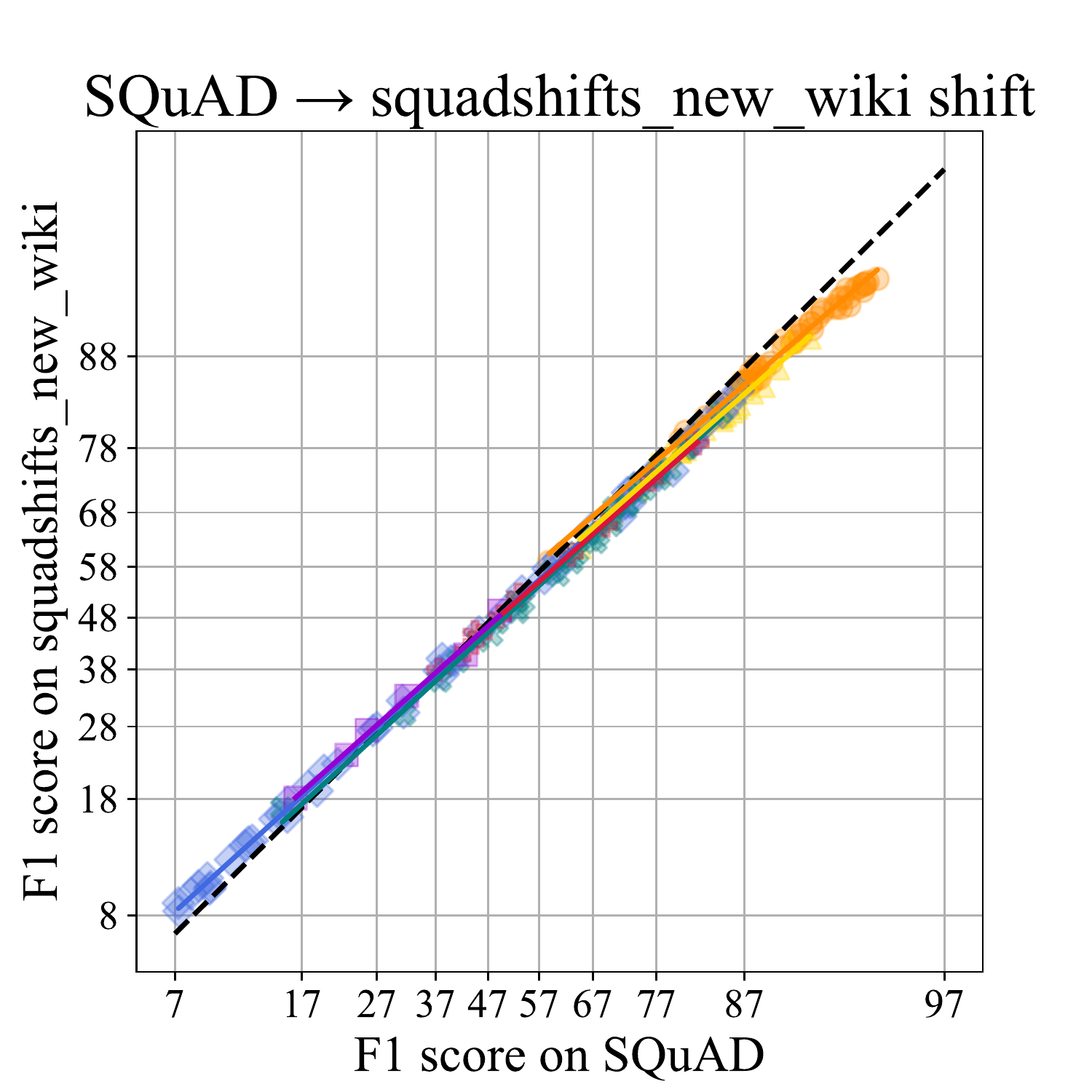}
         \caption{SquadShifts Wikipedia}
          \label{fig:all shifts wiki}
     \end{subfigure}
     \begin{subfigure}[b]{0.32\textwidth}
         \centering
         \includegraphics[width=\textwidth]{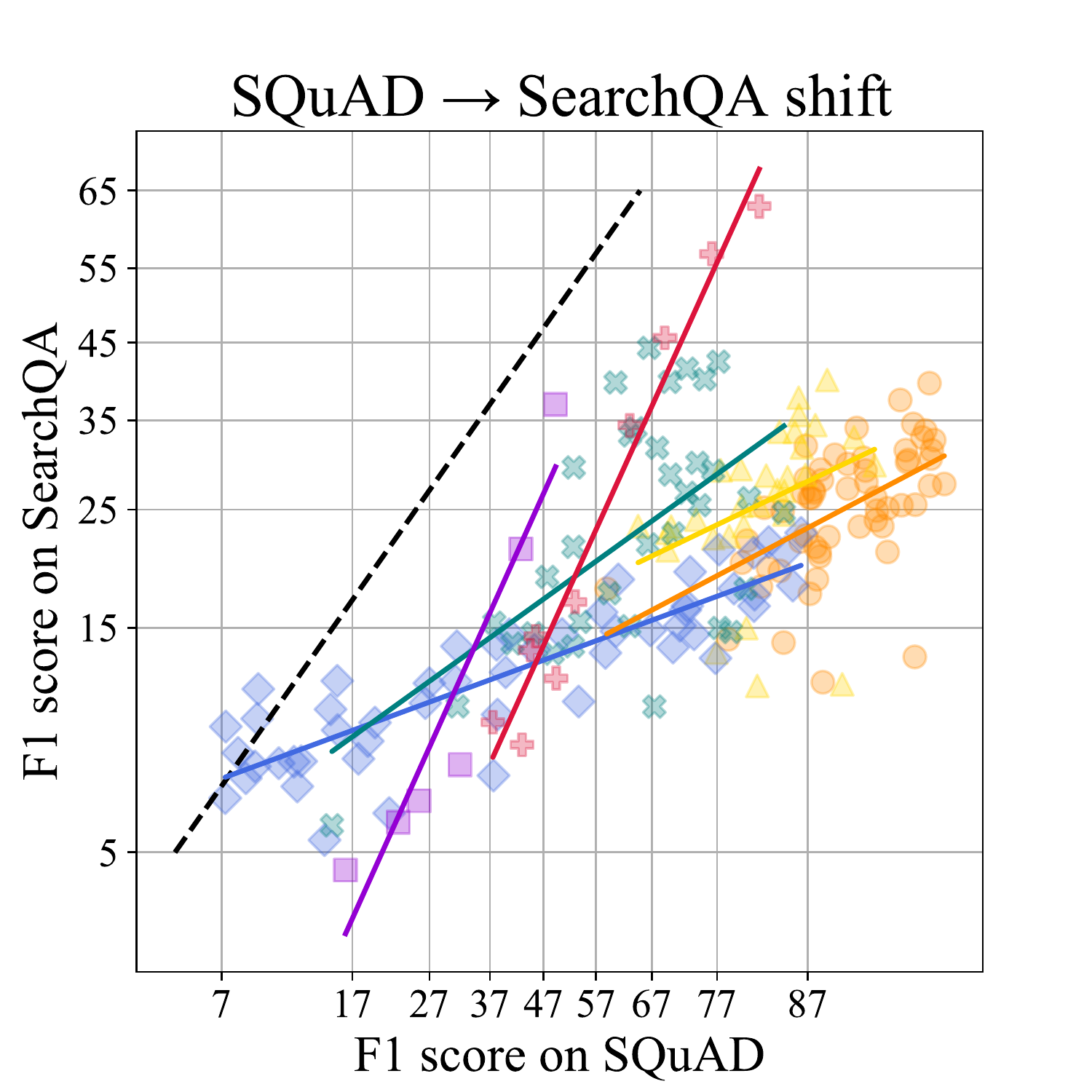}
         \caption{SearchQA}
         \label{fig:all shifts searchqa}
     \end{subfigure}
     \begin{subfigure}[b]{0.32\textwidth}
         \centering
         \includegraphics[width=\textwidth]{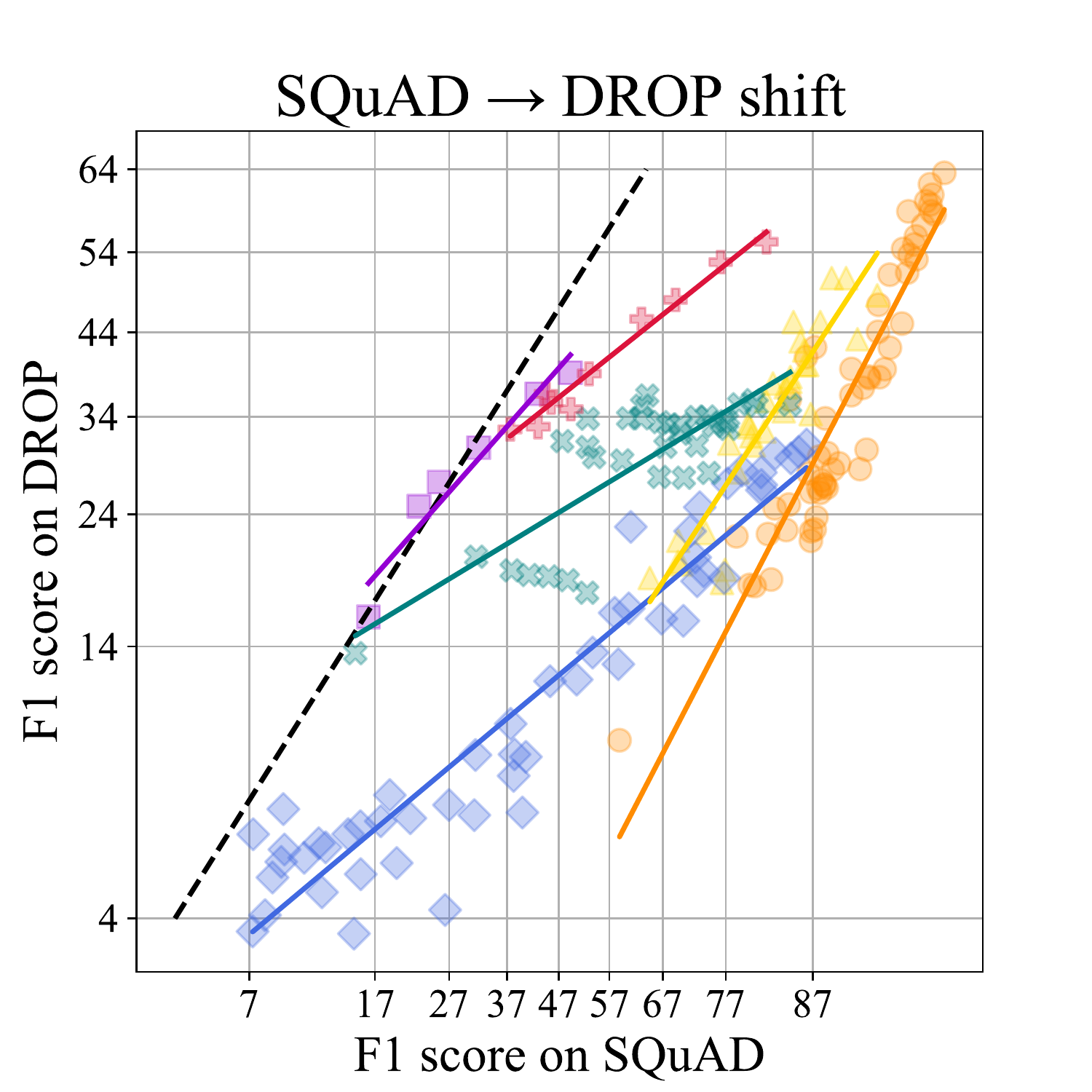}
         \caption{DROP}
         \label{fig:all shifts drop}
     \end{subfigure}
     
     \vspace{0.1cm}
     
     \begin{subfigure}[b]{0.4\textwidth}
         \centering
         \includegraphics[width=\textwidth]{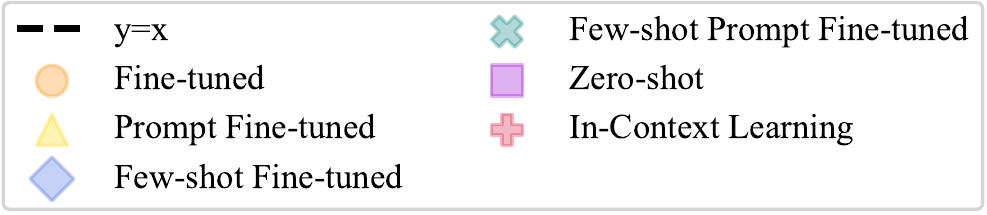}
     \end{subfigure}
     \caption{Instead of averaging over all 15 datasets, we now show logit-scaled plots examining the three distribution shifts individually.
     \textbf{(left)} The SquadShifts Wiki dataset is derived from the same data source (Wikipedia) as SQuAD. As a result, models lie closer to the $y=x$ diagonal than on other distribution shifts.
     \textbf{(middle)} Progress on SQuAD is a weaker indicator for progress on SearchQA for fully fine-tuned models and few-shot fine-tuned models. We find that zero-shot and ICL models are less robust than fine-tuned and few-shot models with the exception of larger language models.
     \textbf{(right)} On the SQuAD${\rightarrow}$DROP distribution shift, we observe that progress beyond $70$ F1 on SQuAD yields quick progress on DROP for fine-tuned models.
     }
     \label{fig:all shifts}
\end{figure*}
\section{Discussion}
\label{sec:discussion}

This section discusses the aforementioned findings. In particular we discuss how the findings compare to analogous studies in vision, and how individual distribution shifts differ from aggregate trends.
\subsection{How do the findings compare to robustness evaluations in vision?}
We observe that the overall robustness trends of question answering models are qualitatively similar to trends identified in image classification~\cite{taori2020measuring,Radford2021LearningTV, miller2021accuracy}. In particular, previous work~\cite{Radford2021LearningTV,wortsman2021robust,pham2021combined} has shown that zero-shot models are more robust than fine-tuned models, which is similar to the trend we observe.
Moreover, additional robustness evaluations~\cite{taori2020measuring} have concluded that fully trained models models with different architectures, pre-training datasets, and robustness enhanced methods do not provide any robustness improvement when evaluated on multiple natural distribution shifts, which is also what we observe.

\subsection{How do individual distribution shifts differ from aggregate trends?}
While we have previously analyzed robustness trends averaged over all distribution shifts, we now examine trends on individual distribution shifts.
For most datasets, we observe qualitatively similar trends as when averaging over all distribution shifts.

One exception is on the SquadShifts New-Wiki dataset, where we find that all models sit very close to the $y=x$ line (Figure~\ref{fig:all shifts wiki}). Since both SQuAD and SquadShifts New-Wiki are collected from Wikipedia, it is perhaps unsurprising that models adapted to SQuAD can generalize to other datasets from the same domain.

Moreover, we observe a piece-wise linear trend when comparing few-shot and fine-tuned models on DROP~\ref{fig:all shifts drop}. By fine-tuning on the entire training set, we improve in-distribution performance, which causes larger gains in DROP performance. Similar patterns of discontinuous improvement have been previously observed by \citet{wei2022emergent}.

Additionally we find that on the SearchQA~\ref{fig:all shifts searchqa} dataset the trendlines are flatter than other distribution shifts for \anas{fine-tuned, prompt fine-tuned and few-shot fine-tuned models} (i.e., increasing ID performance has a smaller impact on OOD performance). In addition, zero-shot and ICL models do not have additional robustness properties. The exception to this is GPT-J and OPT 175B, which continue to outperform other models. \anas{Moreover, few-shot prompt fine-tuned models perform better than other few-shot and fine-tuned models.}

\section{Related work}
Understanding how models behave under conditions that differ from training has been the subject of much attention both in natural language processing \cite[inter alia]{ribeiro-etal-2020-beyond,tu-etal-2020-empirical,hendrycks-etal-2020-pretrained,gardner-etal-2020-evaluating,arora-etal-2021-types,veitch2021counterfactual,goel2021robustness,Miller2020TheEO} and computer vision \cite[inter alia]{pmlr-v97-recht19a,taori2020measuring,miller2021accuracy,koh2021wilds}.
As in \citet{taori2020measuring}, we distinguish between synthetic and natural distribution shifts.
The former includes any artificial perturbations to inputs, including adversarial attacks \cite[inter alia]{szegedy2013intriguing,carlini2017towards,jia2017adversarial,biggio2018wild,wang2019towards,wallace-etal-2019-trick,wallace-etal-2019-universal,tramer2020adaptive,liu2021can,wu-etal-2021-evaluating,chang2021robustness}.
In contrast, the later relates to naturally occurring data, without synthetic or adversarial perturbations.
Our work focuses on natural distribution shifts.

Most similar to our work is that of \citet{yogatama2019learning,talmor2019multiqa,sen2020models,fisch2019mrqa} and \citet{Miller2020TheEO}, who examine the performance of models on multiple question answering datasets.
Our work provides a more comprehensive modeling survey, evaluating a broader set of models, adaptation strategies and datasets.
In contrast to previous work, we evaluate zero-shot inference, in-context learning, few-shot fine-tuning, and parameter-efficient adaptation methods, which have only recently been popularized.

Finally, a variety of methods for improving robustness have been explored by previous work \cite[inter alia]{jiang2019smart,Zhu2020FreeLB,aghajanyan2021better,veitch2021counterfactual,wortsman2021robust}.
Instead of proposing methods to build more robust models, our goal is to empirically examine the landscape of robustness.
As part of this goal, we evaluate robustness enhancing methods, in addition to other adaptation strategies.

Concurrent work by ~\citet{Liu2022AreSN} examines the robustness of few-shot fine-tuned models. They find that these models yield no additional robustness which matches the findings from our evaluation.

\section{Conclusion}
\label{sec:conclusion}
We conduct an extensive evaluation of the robustness of different model and adaptation methods on 15 distribution shifts in question answering. 
Our in-depth analysis suggests several concrete directions for future work: improving the in-distribution performance of ICL methods and understanding \anas{why different few-shot fine-tuning methods yield varied robustness.}
\section{Limitations}
\label{sec:limitations}
\paragraph{Experimenting with different in-distribution datasets.} We choose SQuAD as a representative in-distribution dataset since it is one of the largest and most popular QA datasets. One limitation of SQuAD is that the training set is mainly collected from Wikipedia articles which may not be optimal for building a QA model that generalizes to many domains. Future work could explore the robustness of models trained on datasets from other domains for increased coverage.

\paragraph{Specialized modeling methods.} Our work does not evaluate models with task or data specific components. As an example \citet{andor-etal-2019-giving} improved performance on DROP~\cite{dua-etal-2019-drop} by using arithmetic programs to improve a model's mathematical reasoning. Evaluating the robustness of methods like these are an exciting area for future investigations.

\paragraph{Few-shot GPT evaluations.} Our results indicate that large GPT models fine-tuned on a smaller number of samples are more robust to distribution shifts compared to \anas{other few-shot fine-tuned models that use a prompt or span prediction. However, GPT-2 XL and GPT-Neo, which both have more than one billion parameters, are larger than all few-shot models we evaluate. Future work could examine the impact of architecture on this trend by evaluating other models with more than a billion parameters like T5.}
\paragraph{Multiple fine-tuning runs.} For fine-tuned models we include a single data-point for each model. However, previous work~\cite{phang2018sentence,Dodge2020FineTuningPL} has shown that different data ordering and weight initialization can lead to large variance in model performance.
In Figure~\ref{fig:runs} we evaluate the robustness of RoBERTa Large models fine-tuned with different data ordering and initialization for the span prediction head~\cite{devlin-etal-2019-bert}. We find that on average the robustness of these models does not differ substantially. 
Further investigation into the effect of random seeds on robustness would improve our understanding of the robustness of individual data points.

\subsection*{Acknowledgements}
This work is in part supported by the NSF AI Institute for Foundations of Machine Learning (IFML), Open Philanthropy, Google, and the Allen Institute for AI.
\bibliography{anthology,custom}

\begin{thebibliography}{81}
\expandafter\ifx\csname natexlab\endcsname\relax\def\natexlab#1{#1}\fi

\bibitem[{Aghajanyan et~al.(2021)Aghajanyan, Shrivastava, Gupta, Goyal,
  Zettlemoyer, and Gupta}]{aghajanyan2021better}
Armen Aghajanyan, Akshat Shrivastava, Anchit Gupta, Naman Goyal, Luke
  Zettlemoyer, and Sonal Gupta. 2021.
\newblock \href {https://openreview.net/forum?id=OQ08SN70M1V} {Better
  fine-tuning by reducing representational collapse}.
\newblock In \emph{International Conference on Learning Representations}.

\bibitem[{Andor et~al.(2019)Andor, He, Lee, and
  Pitler}]{andor-etal-2019-giving}
Daniel Andor, Luheng He, Kenton Lee, and Emily Pitler. 2019.
\newblock \href {https://doi.org/10.18653/v1/D19-1609} {Giving {BERT} a
  calculator: Finding operations and arguments with reading comprehension}.
\newblock In \emph{Proceedings of the 2019 Conference on Empirical Methods in
  Natural Language Processing and the 9th International Joint Conference on
  Natural Language Processing (EMNLP-IJCNLP)}, pages 5947--5952, Hong Kong,
  China. Association for Computational Linguistics.

\bibitem[{Andreassen et~al.(2021)Andreassen, Bahri, Neyshabur, and
  Roelofs}]{andreassen2021evolution}
Anders Andreassen, Yasaman Bahri, Behnam Neyshabur, and Rebecca Roelofs. 2021.
\newblock The evolution of out-of-distribution robustness throughout
  fine-tuning.
\newblock \emph{arXiv preprint arXiv:2106.15831}.

\bibitem[{Arora et~al.(2021)Arora, Huang, and He}]{arora-etal-2021-types}
Udit Arora, William Huang, and He~He. 2021.
\newblock \href {https://doi.org/10.18653/v1/2021.emnlp-main.835} {Types of
  out-of-distribution texts and how to detect them}.
\newblock In \emph{Proceedings of the 2021 Conference on Empirical Methods in
  Natural Language Processing}, pages 10687--10701, Online and Punta Cana,
  Dominican Republic. Association for Computational Linguistics.

\bibitem[{Bender and Koller(2020)}]{bender-koller-2020-climbing}
Emily~M. Bender and Alexander Koller. 2020.
\newblock \href {https://doi.org/10.18653/v1/2020.acl-main.463} {Climbing
  towards {NLU}: {On} meaning, form, and understanding in the age of data}.
\newblock In \emph{Proceedings of the 58th Annual Meeting of the Association
  for Computational Linguistics}, pages 5185--5198, Online. Association for
  Computational Linguistics.

\bibitem[{Biggio and Roli(2018)}]{biggio2018wild}
Battista Biggio and Fabio Roli. 2018.
\newblock Wild patterns: Ten years after the rise of adversarial machine
  learning.
\newblock \emph{Pattern Recognition}, 84:317--331.

\bibitem[{Black et~al.(2021)Black, Gao, Wang, Leahy, and Biderman}]{gpt-neo}
Sid Black, Leo Gao, Phil Wang, Connor Leahy, and Stella Biderman. 2021.
\newblock \href {https://doi.org/10.5281/zenodo.5297715} {{GPT-Neo: Large Scale
  Autoregressive Language Modeling with Mesh-Tensorflow}}.
\newblock {If you use this software, please cite it using these metadata.}

\bibitem[{Brown et~al.(2020{\natexlab{a}})Brown, Mann, Ryder, Subbiah, Kaplan,
  Dhariwal, Neelakantan, Shyam, Sastry, Askell, Agarwal, Herbert-Voss, Krueger,
  Henighan, Child, Ramesh, Ziegler, Wu, Winter, Hesse, Chen, Sigler, Litwin,
  Gray, Chess, Clark, Berner, McCandlish, Radford, Sutskever, and
  Amodei}]{NEURIPS2020_1457c0d6}
Tom Brown, Benjamin Mann, Nick Ryder, Melanie Subbiah, Jared~D Kaplan, Prafulla
  Dhariwal, Arvind Neelakantan, Pranav Shyam, Girish Sastry, Amanda Askell,
  Sandhini Agarwal, Ariel Herbert-Voss, Gretchen Krueger, Tom Henighan, Rewon
  Child, Aditya Ramesh, Daniel Ziegler, Jeffrey Wu, Clemens Winter, Chris
  Hesse, Mark Chen, Eric Sigler, Mateusz Litwin, Scott Gray, Benjamin Chess,
  Jack Clark, Christopher Berner, Sam McCandlish, Alec Radford, Ilya Sutskever,
  and Dario Amodei. 2020{\natexlab{a}}.
\newblock \href
  {https://proceedings.neurips.cc/paper/2020/file/1457c0d6bfcb4967418bfb8ac142f64a-Paper.pdf}
  {Language models are few-shot learners}.
\newblock In \emph{Advances in Neural Information Processing Systems},
  volume~33, pages 1877--1901. Curran Associates, Inc.

\bibitem[{Brown et~al.(2020{\natexlab{b}})Brown, Mann, Ryder, Subbiah, Kaplan,
  Dhariwal, Neelakantan, Shyam, Sastry, Askell et~al.}]{brown2020language}
Tom Brown, Benjamin Mann, Nick Ryder, Melanie Subbiah, Jared~D Kaplan, Prafulla
  Dhariwal, Arvind Neelakantan, Pranav Shyam, Girish Sastry, Amanda Askell,
  et~al. 2020{\natexlab{b}}.
\newblock Language models are few-shot learners.
\newblock \emph{Advances in neural information processing systems},
  33:1877--1901.

\bibitem[{Carlini and Wagner(2017)}]{carlini2017towards}
Nicholas Carlini and David Wagner. 2017.
\newblock Towards evaluating the robustness of neural networks.
\newblock In \emph{2017 ieee symposium on security and privacy (sp)}, pages
  39--57. Ieee.

\bibitem[{Chada and Natarajan(2021)}]{chada-natarajan-2021-fewshotqa}
Rakesh Chada and Pradeep Natarajan. 2021.
\newblock \href {https://doi.org/10.18653/v1/2021.emnlp-main.491}
  {{F}ewshot{QA}: A simple framework for few-shot learning of question
  answering tasks using pre-trained text-to-text models}.
\newblock In \emph{Proceedings of the 2021 Conference on Empirical Methods in
  Natural Language Processing}, pages 6081--6090, Online and Punta Cana,
  Dominican Republic. Association for Computational Linguistics.

\bibitem[{Chang et~al.(2021)Chang, He, Jia, and Singh}]{chang2021robustness}
Kai-Wei Chang, He~He, Robin Jia, and Sameer Singh. 2021.
\newblock Robustness and adversarial examples in natural language processing.
\newblock In \emph{Proceedings of the 2021 Conference on Empirical Methods in
  Natural Language Processing: Tutorial Abstracts}, pages 22--26.

\bibitem[{Chowdhery et~al.(2022)Chowdhery, Narang, Devlin, Bosma, Mishra,
  Roberts, Barham, Chung, Sutton, Gehrmann et~al.}]{chowdhery2022palm}
Aakanksha Chowdhery, Sharan Narang, Jacob Devlin, Maarten Bosma, Gaurav Mishra,
  Adam Roberts, Paul Barham, Hyung~Won Chung, Charles Sutton, Sebastian
  Gehrmann, et~al. 2022.
\newblock Palm: Scaling language modeling with pathways.
\newblock \emph{arXiv preprint arXiv:2204.02311}.

\bibitem[{Devlin et~al.(2019)Devlin, Chang, Lee, and
  Toutanova}]{devlin-etal-2019-bert}
Jacob Devlin, Ming-Wei Chang, Kenton Lee, and Kristina Toutanova. 2019.
\newblock \href {https://doi.org/10.18653/v1/N19-1423} {{BERT}: Pre-training of
  deep bidirectional transformers for language understanding}.
\newblock In \emph{Proceedings of the 2019 Conference of the North {A}merican
  Chapter of the Association for Computational Linguistics: Human Language
  Technologies, Volume 1 (Long and Short Papers)}, pages 4171--4186,
  Minneapolis, Minnesota. Association for Computational Linguistics.

\bibitem[{Dodge et~al.(2020)Dodge, Ilharco, Schwartz, Farhadi, Hajishirzi, and
  Smith}]{Dodge2020FineTuningPL}
Jesse Dodge, Gabriel Ilharco, Roy Schwartz, Ali Farhadi, Hannaneh Hajishirzi,
  and Noah~A. Smith. 2020.
\newblock Fine-tuning pretrained language models: Weight initializations, data
  orders, and early stopping.
\newblock \emph{ArXiv}, abs/2002.06305.

\bibitem[{Dua et~al.(2019)Dua, Wang, Dasigi, Stanovsky, Singh, and
  Gardner}]{dua-etal-2019-drop}
Dheeru Dua, Yizhong Wang, Pradeep Dasigi, Gabriel Stanovsky, Sameer Singh, and
  Matt Gardner. 2019.
\newblock \href {https://doi.org/10.18653/v1/N19-1246} {{DROP}: A reading
  comprehension benchmark requiring discrete reasoning over paragraphs}.
\newblock In \emph{Proceedings of the 2019 Conference of the North {A}merican
  Chapter of the Association for Computational Linguistics: Human Language
  Technologies, Volume 1 (Long and Short Papers)}, pages 2368--2378,
  Minneapolis, Minnesota. Association for Computational Linguistics.

\bibitem[{Dunn et~al.(2017)Dunn, Sagun, Higgins, G{\"u}ney, Cirik, and
  Cho}]{Dunn2017SearchQAAN}
Matthew Dunn, Levent Sagun, Mike Higgins, V.~Ugur G{\"u}ney, Volkan Cirik, and
  Kyunghyun Cho. 2017.
\newblock Searchqa: A new q\&a dataset augmented with context from a search
  engine.
\newblock \emph{ArXiv}, abs/1704.05179.

\bibitem[{Fisch et~al.(2019)Fisch, Talmor, Jia, Seo, Choi, and
  Chen}]{fisch2019mrqa}
Adam Fisch, Alon Talmor, Robin Jia, Minjoon Seo, Eunsol Choi, and Danqi Chen.
  2019.
\newblock {MRQA} 2019 shared task: Evaluating generalization in reading
  comprehension.
\newblock In \emph{Proceedings of 2nd Machine Reading for Reading Comprehension
  (MRQA) Workshop at EMNLP}.

\bibitem[{Gardner et~al.(2020)Gardner, Artzi, Basmov, Berant, Bogin, Chen,
  Dasigi, Dua, Elazar, Gottumukkala, Gupta, Hajishirzi, Ilharco, Khashabi, Lin,
  Liu, Liu, Mulcaire, Ning, Singh, Smith, Subramanian, Tsarfaty, Wallace,
  Zhang, and Zhou}]{gardner-etal-2020-evaluating}
Matt Gardner, Yoav Artzi, Victoria Basmov, Jonathan Berant, Ben Bogin, Sihao
  Chen, Pradeep Dasigi, Dheeru Dua, Yanai Elazar, Ananth Gottumukkala, Nitish
  Gupta, Hannaneh Hajishirzi, Gabriel Ilharco, Daniel Khashabi, Kevin Lin,
  Jiangming Liu, Nelson~F. Liu, Phoebe Mulcaire, Qiang Ning, Sameer Singh,
  Noah~A. Smith, Sanjay Subramanian, Reut Tsarfaty, Eric Wallace, Ally Zhang,
  and Ben Zhou. 2020.
\newblock \href {https://doi.org/10.18653/v1/2020.findings-emnlp.117}
  {Evaluating models{'} local decision boundaries via contrast sets}.
\newblock In \emph{Findings of the Association for Computational Linguistics:
  EMNLP 2020}, pages 1307--1323, Online. Association for Computational
  Linguistics.

\bibitem[{Goel et~al.(2021)Goel, Rajani, Vig, Tan, Wu, Zheng, Xiong, Bansal,
  and R{\'e}}]{goel2021robustness}
Karan Goel, Nazneen Rajani, Jesse Vig, Samson Tan, Jason Wu, Stephan Zheng,
  Caiming Xiong, Mohit Bansal, and Christopher R{\'e}. 2021.
\newblock Robustness gym: Unifying the nlp evaluation landscape.
\newblock \emph{arXiv preprint arXiv:2101.04840}.

\bibitem[{Hendrycks et~al.(2020)Hendrycks, Liu, Wallace, Dziedzic, Krishnan,
  and Song}]{hendrycks-etal-2020-pretrained}
Dan Hendrycks, Xiaoyuan Liu, Eric Wallace, Adam Dziedzic, Rishabh Krishnan, and
  Dawn Song. 2020.
\newblock \href {https://doi.org/10.18653/v1/2020.acl-main.244} {Pretrained
  transformers improve out-of-distribution robustness}.
\newblock In \emph{Proceedings of the 58th Annual Meeting of the Association
  for Computational Linguistics}, pages 2744--2751, Online. Association for
  Computational Linguistics.

\bibitem[{Hoffmann et~al.(2022)Hoffmann, Borgeaud, Mensch, Buchatskaya, Cai,
  Rutherford, Casas, Hendricks, Welbl, Clark et~al.}]{hoffmann2022training}
Jordan Hoffmann, Sebastian Borgeaud, Arthur Mensch, Elena Buchatskaya, Trevor
  Cai, Eliza Rutherford, Diego de~Las Casas, Lisa~Anne Hendricks, Johannes
  Welbl, Aidan Clark, et~al. 2022.
\newblock Training compute-optimal large language models.
\newblock \emph{arXiv preprint arXiv:2203.15556}.

\bibitem[{Houlsby et~al.(2019)Houlsby, Giurgiu, Jastrzebski, Morrone,
  de~Laroussilhe, Gesmundo, Attariyan, and
  Gelly}]{Houlsby2019ParameterEfficientTL}
Neil Houlsby, Andrei Giurgiu, Stanislaw Jastrzebski, Bruna Morrone, Quentin
  de~Laroussilhe, Andrea Gesmundo, Mona Attariyan, and Sylvain Gelly. 2019.
\newblock Parameter-efficient transfer learning for nlp.
\newblock In \emph{ICML}.

\bibitem[{Hu et~al.(2021)Hu, Shen, Wallis, Allen-Zhu, Li, Wang, Wang, and
  Chen}]{hu2021lora}
Edward~J Hu, Yelong Shen, Phillip Wallis, Zeyuan Allen-Zhu, Yuanzhi Li, Shean
  Wang, Lu~Wang, and Weizhu Chen. 2021.
\newblock Lora: Low-rank adaptation of large language models.
\newblock \emph{arXiv preprint arXiv:2106.09685}.

\bibitem[{Jia and Liang(2017)}]{jia2017adversarial}
Robin Jia and Percy Liang. 2017.
\newblock Adversarial examples for evaluating reading comprehension systems.
\newblock \emph{arXiv preprint arXiv:1707.07328}.

\bibitem[{Jiang et~al.(2019)Jiang, He, Chen, Liu, Gao, and
  Zhao}]{jiang2019smart}
Haoming Jiang, Pengcheng He, Weizhu Chen, Xiaodong Liu, Jianfeng Gao, and Tuo
  Zhao. 2019.
\newblock Smart: Robust and efficient fine-tuning for pre-trained natural
  language models through principled regularized optimization.
\newblock \emph{arXiv preprint arXiv:1911.03437}.

\bibitem[{Joshi et~al.(2020)Joshi, Chen, Liu, Weld, Zettlemoyer, and
  Levy}]{joshi-etal-2020-spanbert}
Mandar Joshi, Danqi Chen, Yinhan Liu, Daniel~S. Weld, Luke Zettlemoyer, and
  Omer Levy. 2020.
\newblock \href {https://doi.org/10.1162/tacl_a_00300} {{S}pan{BERT}: Improving
  pre-training by representing and predicting spans}.
\newblock \emph{Transactions of the Association for Computational Linguistics},
  8:64--77.

\bibitem[{Joshi et~al.(2017)Joshi, Choi, Weld, and
  Zettlemoyer}]{joshi-etal-2017-triviaqa}
Mandar Joshi, Eunsol Choi, Daniel Weld, and Luke Zettlemoyer. 2017.
\newblock \href {https://doi.org/10.18653/v1/P17-1147} {{T}rivia{QA}: A large
  scale distantly supervised challenge dataset for reading comprehension}.
\newblock In \emph{Proceedings of the 55th Annual Meeting of the Association
  for Computational Linguistics (Volume 1: Long Papers)}, pages 1601--1611,
  Vancouver, Canada. Association for Computational Linguistics.

\bibitem[{Kembhavi et~al.(2017)Kembhavi, Seo, Schwenk, Choi, Farhadi, and
  Hajishirzi}]{Kembhavi_2017_CVPR}
Aniruddha Kembhavi, Minjoon Seo, Dustin Schwenk, Jonghyun Choi, Ali Farhadi,
  and Hannaneh Hajishirzi. 2017.
\newblock Are you smarter than a sixth grader? textbook question answering for
  multimodal machine comprehension.
\newblock In \emph{Proceedings of the IEEE Conference on Computer Vision and
  Pattern Recognition (CVPR)}.

\bibitem[{Koh et~al.(2021)Koh, Sagawa, Marklund, Xie, Zhang, Balsubramani, Hu,
  Yasunaga, Phillips, Gao et~al.}]{koh2021wilds}
Pang~Wei Koh, Shiori Sagawa, Henrik Marklund, Sang~Michael Xie, Marvin Zhang,
  Akshay Balsubramani, Weihua Hu, Michihiro Yasunaga, Richard~Lanas Phillips,
  Irena Gao, et~al. 2021.
\newblock Wilds: A benchmark of in-the-wild distribution shifts.
\newblock In \emph{International Conference on Machine Learning}, pages
  5637--5664. PMLR.

\bibitem[{Kwiatkowski et~al.(2019)Kwiatkowski, Palomaki, Redfield, Collins,
  Parikh, Alberti, Epstein, Polosukhin, Devlin, Lee, Toutanova, Jones, Kelcey,
  Chang, Dai, Uszkoreit, Le, and Petrov}]{kwiatkowski-etal-2019-natural}
Tom Kwiatkowski, Jennimaria Palomaki, Olivia Redfield, Michael Collins, Ankur
  Parikh, Chris Alberti, Danielle Epstein, Illia Polosukhin, Jacob Devlin,
  Kenton Lee, Kristina Toutanova, Llion Jones, Matthew Kelcey, Ming-Wei Chang,
  Andrew~M. Dai, Jakob Uszkoreit, Quoc Le, and Slav Petrov. 2019.
\newblock \href {https://doi.org/10.1162/tacl_a_00276} {Natural questions: A
  benchmark for question answering research}.
\newblock \emph{Transactions of the Association for Computational Linguistics},
  7:452--466.

\bibitem[{Lai et~al.(2017)Lai, Xie, Liu, Yang, and Hovy}]{lai-etal-2017-race}
Guokun Lai, Qizhe Xie, Hanxiao Liu, Yiming Yang, and Eduard Hovy. 2017.
\newblock \href {https://doi.org/10.18653/v1/D17-1082} {{RACE}: Large-scale
  {R}e{A}ding comprehension dataset from examinations}.
\newblock In \emph{Proceedings of the 2017 Conference on Empirical Methods in
  Natural Language Processing}, pages 785--794, Copenhagen, Denmark.
  Association for Computational Linguistics.

\bibitem[{Lan et~al.(2020)Lan, Chen, Goodman, Gimpel, Sharma, and
  Soricut}]{Lan2020ALBERT}
Zhenzhong Lan, Mingda Chen, Sebastian Goodman, Kevin Gimpel, Piyush Sharma, and
  Radu Soricut. 2020.
\newblock Albert: A lite bert for self-supervised learning of language
  representations.
\newblock In \emph{International Conference on Learning Representations}.

\bibitem[{Lester et~al.(2021)Lester, Al-Rfou, and
  Constant}]{lester-etal-2021-power}
Brian Lester, Rami Al-Rfou, and Noah Constant. 2021.
\newblock \href {https://doi.org/10.18653/v1/2021.emnlp-main.243} {The power of
  scale for parameter-efficient prompt tuning}.
\newblock In \emph{Proceedings of the 2021 Conference on Empirical Methods in
  Natural Language Processing}, pages 3045--3059, Online and Punta Cana,
  Dominican Republic. Association for Computational Linguistics.

\bibitem[{Levy et~al.(2017)Levy, Seo, Choi, and
  Zettlemoyer}]{levy-etal-2017-zero}
Omer Levy, Minjoon Seo, Eunsol Choi, and Luke Zettlemoyer. 2017.
\newblock \href {https://doi.org/10.18653/v1/K17-1034} {Zero-shot relation
  extraction via reading comprehension}.
\newblock In \emph{Proceedings of the 21st Conference on Computational Natural
  Language Learning ({C}o{NLL} 2017)}, pages 333--342, Vancouver, Canada.
  Association for Computational Linguistics.

\bibitem[{Lewis et~al.(2020)Lewis, Liu, Goyal, Ghazvininejad, Mohamed, Levy,
  Stoyanov, and Zettlemoyer}]{lewis-etal-2020-bart}
Mike Lewis, Yinhan Liu, Naman Goyal, Marjan Ghazvininejad, Abdelrahman Mohamed,
  Omer Levy, Veselin Stoyanov, and Luke Zettlemoyer. 2020.
\newblock \href {https://doi.org/10.18653/v1/2020.acl-main.703} {{BART}:
  Denoising sequence-to-sequence pre-training for natural language generation,
  translation, and comprehension}.
\newblock In \emph{Proceedings of the 58th Annual Meeting of the Association
  for Computational Linguistics}, pages 7871--7880, Online. Association for
  Computational Linguistics.

\bibitem[{Li and Liang(2021)}]{li-liang-2021-prefix}
Xiang~Lisa Li and Percy Liang. 2021.
\newblock \href {https://doi.org/10.18653/v1/2021.acl-long.353} {Prefix-tuning:
  Optimizing continuous prompts for generation}.
\newblock In \emph{Proceedings of the 59th Annual Meeting of the Association
  for Computational Linguistics and the 11th International Joint Conference on
  Natural Language Processing (Volume 1: Long Papers)}, pages 4582--4597,
  Online. Association for Computational Linguistics.

\bibitem[{Liu et~al.(2022)Liu, Kumar, Liang, and Jia}]{Liu2022AreSN}
Nelson~F. Liu, Ananya Kumar, Percy Liang, and Robin Jia. 2022.
\newblock Are sample-efficient nlp models more robust?
\newblock \emph{ArXiv}, abs/2210.06456.

\bibitem[{Liu et~al.(2021)Liu, Lee, Jia, and Liang}]{liu2021can}
Nelson~F Liu, Tony Lee, Robin Jia, and Percy Liang. 2021.
\newblock Can small and synthetic benchmarks drive modeling innovation? a
  retrospective study of question answering modeling approaches.
\newblock \emph{arXiv preprint arXiv:2102.01065}.

\bibitem[{Liu et~al.(2019)Liu, Ott, Goyal, Du, Joshi, Chen, Levy, Lewis,
  Zettlemoyer, and Stoyanov}]{liu2019roberta}
Yinhan Liu, Myle Ott, Naman Goyal, Jingfei Du, Mandar Joshi, Danqi Chen, Omer
  Levy, Mike Lewis, Luke Zettlemoyer, and Veselin Stoyanov. 2019.
\newblock Roberta: A robustly optimized bert pretraining approach.
\newblock \emph{arXiv preprint arXiv:1907.11692}.

\bibitem[{Luu et~al.(2021)Luu, Khashabi, Gururangan, Mandyam, and
  Smith}]{luu2021time}
Kelvin Luu, Daniel Khashabi, Suchin Gururangan, Karishma Mandyam, and Noah~A
  Smith. 2021.
\newblock Time waits for no one! analysis and challenges of temporal
  misalignment.
\newblock \emph{arXiv preprint arXiv:2111.07408}.

\bibitem[{Miller et~al.(2020)Miller, Krauth, Recht, and
  Schmidt}]{Miller2020TheEO}
John Miller, Karl Krauth, Benjamin Recht, and Ludwig Schmidt. 2020.
\newblock The effect of natural distribution shift on question answering
  models.
\newblock In \emph{ICML}.

\bibitem[{Miller et~al.(2021{\natexlab{a}})Miller, Taori, Raghunathan, Sagawa,
  Koh, Shankar, Liang, Carmon, and Schmidt}]{miller2021accuracy}
John~P Miller, Rohan Taori, Aditi Raghunathan, Shiori Sagawa, Pang~Wei Koh,
  Vaishaal Shankar, Percy Liang, Yair Carmon, and Ludwig Schmidt.
  2021{\natexlab{a}}.
\newblock Accuracy on the line: on the strong correlation between
  out-of-distribution and in-distribution generalization.
\newblock In \emph{International Conference on Machine Learning}, pages
  7721--7735. PMLR.

\bibitem[{Miller et~al.(2021{\natexlab{b}})Miller, Taori, Raghunathan, Sagawa,
  Koh, Shankar, Liang, Carmon, and Schmidt}]{pmlr-v139-miller21b}
John~P Miller, Rohan Taori, Aditi Raghunathan, Shiori Sagawa, Pang~Wei Koh,
  Vaishaal Shankar, Percy Liang, Yair Carmon, and Ludwig Schmidt.
  2021{\natexlab{b}}.
\newblock \href {https://proceedings.mlr.press/v139/miller21b.html} {Accuracy
  on the line: on the strong correlation between out-of-distribution and
  in-distribution generalization}.
\newblock In \emph{Proceedings of the 38th International Conference on Machine
  Learning}, volume 139 of \emph{Proceedings of Machine Learning Research},
  pages 7721--7735. PMLR.

\bibitem[{Pfeiffer et~al.(2021)Pfeiffer, Kamath, R{\"u}ckl{\'e}, Cho, and
  Gurevych}]{pfeiffer-etal-2021-adapterfusion}
Jonas Pfeiffer, Aishwarya Kamath, Andreas R{\"u}ckl{\'e}, Kyunghyun Cho, and
  Iryna Gurevych. 2021.
\newblock \href {https://doi.org/10.18653/v1/2021.eacl-main.39}
  {{A}dapter{F}usion: Non-destructive task composition for transfer learning}.
\newblock In \emph{Proceedings of the 16th Conference of the European Chapter
  of the Association for Computational Linguistics: Main Volume}, pages
  487--503, Online. Association for Computational Linguistics.

\bibitem[{Pfeiffer et~al.(2020)Pfeiffer, R{\"u}ckl{\'e}, Poth, Kamath,
  Vuli{\'c}, Ruder, Cho, and Gurevych}]{pfeiffer-etal-2020-adapterhub}
Jonas Pfeiffer, Andreas R{\"u}ckl{\'e}, Clifton Poth, Aishwarya Kamath, Ivan
  Vuli{\'c}, Sebastian Ruder, Kyunghyun Cho, and Iryna Gurevych. 2020.
\newblock \href {https://doi.org/10.18653/v1/2020.emnlp-demos.7}
  {{A}dapter{H}ub: A framework for adapting transformers}.
\newblock In \emph{Proceedings of the 2020 Conference on Empirical Methods in
  Natural Language Processing: System Demonstrations}, pages 46--54, Online.
  Association for Computational Linguistics.

\bibitem[{Pham et~al.(2021)Pham, Dai, Ghiasi, Liu, Yu, Luong, Tan, and
  Le}]{pham2021combined}
Hieu Pham, Zihang Dai, Golnaz Ghiasi, Hanxiao Liu, Adams~Wei Yu, Minh-Thang
  Luong, Mingxing Tan, and Quoc~V Le. 2021.
\newblock Combined scaling for zero-shot transfer learning.
\newblock \emph{arXiv preprint arXiv:2111.10050}.

\bibitem[{Phang et~al.(2018)Phang, F{\'e}vry, and Bowman}]{phang2018sentence}
Jason Phang, Thibault F{\'e}vry, and Samuel~R Bowman. 2018.
\newblock Sentence encoders on stilts: Supplementary training on intermediate
  labeled-data tasks.
\newblock \emph{arXiv preprint arXiv:1811.01088}.

\bibitem[{Radford et~al.(2021)Radford, Kim, Hallacy, Ramesh, Goh, Agarwal,
  Sastry, Askell, Mishkin, Clark, Krueger, and
  Sutskever}]{Radford2021LearningTV}
Alec Radford, Jong~Wook Kim, Chris Hallacy, Aditya Ramesh, Gabriel Goh,
  Sandhini Agarwal, Girish Sastry, Amanda Askell, Pamela Mishkin, Jack Clark,
  Gretchen Krueger, and Ilya Sutskever. 2021.
\newblock Learning transferable visual models from natural language
  supervision.
\newblock In \emph{ICML}.

\bibitem[{Radford et~al.(2019)Radford, Wu, Child, Luan, Amodei, and
  Sutskever}]{Radford2019LanguageMA}
Alec Radford, Jeff Wu, Rewon Child, David Luan, Dario Amodei, and Ilya
  Sutskever. 2019.
\newblock Language models are unsupervised multitask learners.

\bibitem[{Raffel et~al.(2019)Raffel, Shazeer, Roberts, Lee, Narang, Matena,
  Zhou, Li, and Liu}]{raffel2019exploring}
Colin Raffel, Noam Shazeer, Adam Roberts, Katherine Lee, Sharan Narang, Michael
  Matena, Yanqi Zhou, Wei Li, and Peter~J Liu. 2019.
\newblock Exploring the limits of transfer learning with a unified text-to-text
  transformer.
\newblock \emph{arXiv preprint arXiv:1910.10683}.

\bibitem[{Rajpurkar et~al.(2016)Rajpurkar, Zhang, Lopyrev, and
  Liang}]{rajpurkar-etal-2016-squad}
Pranav Rajpurkar, Jian Zhang, Konstantin Lopyrev, and Percy Liang. 2016.
\newblock \href {https://doi.org/10.18653/v1/D16-1264} {{SQ}u{AD}: 100,000+
  questions for machine comprehension of text}.
\newblock In \emph{Proceedings of the 2016 Conference on Empirical Methods in
  Natural Language Processing}, pages 2383--2392, Austin, Texas. Association
  for Computational Linguistics.

\bibitem[{Ram et~al.(2021{\natexlab{a}})Ram, Kirstain, Berant, Globerson, and
  Levy}]{Ram2021FewShotQA}
Ori Ram, Yuval Kirstain, Jonathan Berant, Amir Globerson, and Omer Levy.
  2021{\natexlab{a}}.
\newblock Few-shot question answering by pretraining span selection.
\newblock In \emph{ACL}.

\bibitem[{Ram et~al.(2021{\natexlab{b}})Ram, Kirstain, Berant, Globerson, and
  Levy}]{ram-etal-2021-shot}
Ori Ram, Yuval Kirstain, Jonathan Berant, Amir Globerson, and Omer Levy.
  2021{\natexlab{b}}.
\newblock \href {https://doi.org/10.18653/v1/2021.acl-long.239} {Few-shot
  question answering by pretraining span selection}.
\newblock In \emph{Proceedings of the 59th Annual Meeting of the Association
  for Computational Linguistics and the 11th International Joint Conference on
  Natural Language Processing (Volume 1: Long Papers)}, pages 3066--3079,
  Online. Association for Computational Linguistics.

\bibitem[{Recht et~al.(2019)Recht, Roelofs, Schmidt, and
  Shankar}]{pmlr-v97-recht19a}
Benjamin Recht, Rebecca Roelofs, Ludwig Schmidt, and Vaishaal Shankar. 2019.
\newblock \href {https://proceedings.mlr.press/v97/recht19a.html} {Do
  {I}mage{N}et classifiers generalize to {I}mage{N}et?}
\newblock In \emph{Proceedings of the 36th International Conference on Machine
  Learning}, volume~97 of \emph{Proceedings of Machine Learning Research},
  pages 5389--5400. PMLR.

\bibitem[{Ribeiro and Lundberg(2022)}]{ribeiro-lundberg-2022-adaptive}
Marco~Tulio Ribeiro and Scott Lundberg. 2022.
\newblock \href {https://aclanthology.org/2022.acl-long.230} {Adaptive testing
  and debugging of {NLP} models}.
\newblock In \emph{Proceedings of the 60th Annual Meeting of the Association
  for Computational Linguistics (Volume 1: Long Papers)}, pages 3253--3267,
  Dublin, Ireland. Association for Computational Linguistics.

\bibitem[{Ribeiro et~al.(2020)Ribeiro, Wu, Guestrin, and
  Singh}]{ribeiro-etal-2020-beyond}
Marco~Tulio Ribeiro, Tongshuang Wu, Carlos Guestrin, and Sameer Singh. 2020.
\newblock \href {https://doi.org/10.18653/v1/2020.acl-main.442} {Beyond
  accuracy: Behavioral testing of {NLP} models with {C}heck{L}ist}.
\newblock In \emph{Proceedings of the 58th Annual Meeting of the Association
  for Computational Linguistics}, pages 4902--4912, Online. Association for
  Computational Linguistics.

\bibitem[{Saha et~al.(2018)Saha, Aralikatte, Khapra, and
  Sankaranarayanan}]{Saha2018DuoRCTC}
Amrita Saha, Rahul Aralikatte, Mitesh~M. Khapra, and Karthik Sankaranarayanan.
  2018.
\newblock Duorc: Towards complex language understanding with paraphrased
  reading comprehension.
\newblock In \emph{ACL}.

\bibitem[{Sen and Saffari(2020)}]{sen2020models}
Priyanka Sen and Amir Saffari. 2020.
\newblock What do models learn from question answering datasets?
\newblock \emph{arXiv preprint arXiv:2004.03490}.

\bibitem[{Szegedy et~al.(2013)Szegedy, Zaremba, Sutskever, Bruna, Erhan,
  Goodfellow, and Fergus}]{szegedy2013intriguing}
Christian Szegedy, Wojciech Zaremba, Ilya Sutskever, Joan Bruna, Dumitru Erhan,
  Ian Goodfellow, and Rob Fergus. 2013.
\newblock Intriguing properties of neural networks.
\newblock \emph{arXiv preprint arXiv:1312.6199}.

\bibitem[{Talmor and Berant(2019)}]{talmor2019multiqa}
Alon Talmor and Jonathan Berant. 2019.
\newblock Multiqa: An empirical investigation of generalization and transfer in
  reading comprehension.
\newblock \emph{arXiv preprint arXiv:1905.13453}.

\bibitem[{Taori et~al.(2020)Taori, Dave, Shankar, Carlini, Recht, and
  Schmidt}]{taori2020measuring}
Rohan Taori, Achal Dave, Vaishaal Shankar, Nicholas Carlini, Benjamin Recht,
  and Ludwig Schmidt. 2020.
\newblock Measuring robustness to natural distribution shifts in image
  classification.
\newblock \emph{Advances in Neural Information Processing Systems},
  33:18583--18599.

\bibitem[{Tramer et~al.(2020)Tramer, Carlini, Brendel, and
  Madry}]{tramer2020adaptive}
Florian Tramer, Nicholas Carlini, Wieland Brendel, and Aleksander Madry. 2020.
\newblock On adaptive attacks to adversarial example defenses.
\newblock \emph{Advances in Neural Information Processing Systems},
  33:1633--1645.

\bibitem[{Trischler et~al.(2017)Trischler, Wang, Yuan, Harris, Sordoni,
  Bachman, and Suleman}]{trischler-etal-2017-newsqa}
Adam Trischler, Tong Wang, Xingdi Yuan, Justin Harris, Alessandro Sordoni,
  Philip Bachman, and Kaheer Suleman. 2017.
\newblock \href {https://doi.org/10.18653/v1/W17-2623} {{N}ews{QA}: A machine
  comprehension dataset}.
\newblock In \emph{Proceedings of the 2nd Workshop on Representation Learning
  for {NLP}}, pages 191--200, Vancouver, Canada. Association for Computational
  Linguistics.

\bibitem[{Tsatsaronis et~al.(2015)Tsatsaronis, Balikas, Malakasiotis, Partalas,
  Zschunke, Alvers, Weißenborn, Krithara, Petridis, Polychronopoulos,
  Almirantis, Pavlopoulos, Baskiotis, Gallinari, Artieres, Ngonga~Ngomo, Heino,
  Gaussier, Barrio-Alvers, and Paliouras}]{TsatsaronisBioasq}
George Tsatsaronis, Georgios Balikas, Prodromos Malakasiotis, Ioannis Partalas,
  Matthias Zschunke, Michael Alvers, Dirk Weißenborn, Anastasia Krithara,
  Sergios Petridis, Dimitris Polychronopoulos, Yannis Almirantis, John
  Pavlopoulos, Nicolas Baskiotis, Patrick Gallinari, Thierry Artieres,
  Axel-Cyrille Ngonga~Ngomo, Norman Heino, Eric Gaussier, Liliana
  Barrio-Alvers, and Georgios Paliouras. 2015.
\newblock \href {https://doi.org/10.1186/s12859-015-0564-6} {An overview of the
  bioasq large-scale biomedical semantic indexing and question answering
  competition}.
\newblock \emph{BMC Bioinformatics}, 16:138.

\bibitem[{Tu et~al.(2020)Tu, Lalwani, Gella, and He}]{tu-etal-2020-empirical}
Lifu Tu, Garima Lalwani, Spandana Gella, and He~He. 2020.
\newblock \href {https://doi.org/10.1162/tacl_a_00335} {An empirical study on
  robustness to spurious correlations using pre-trained language models}.
\newblock \emph{Transactions of the Association for Computational Linguistics},
  8:621--633.

\bibitem[{Veitch et~al.(2021)Veitch, D'Amour, Yadlowsky, and
  Eisenstein}]{veitch2021counterfactual}
Victor Veitch, Alexander D'Amour, Steve Yadlowsky, and Jacob Eisenstein. 2021.
\newblock Counterfactual invariance to spurious correlations: Why and how to
  pass stress tests.
\newblock Advances in Neural Information Processing Systems.

\bibitem[{Wallace et~al.(2019{\natexlab{a}})Wallace, Feng, Kandpal, Gardner,
  and Singh}]{wallace-etal-2019-universal}
Eric Wallace, Shi Feng, Nikhil Kandpal, Matt Gardner, and Sameer Singh.
  2019{\natexlab{a}}.
\newblock \href {https://doi.org/10.18653/v1/D19-1221} {Universal adversarial
  triggers for attacking and analyzing {NLP}}.
\newblock In \emph{Proceedings of the 2019 Conference on Empirical Methods in
  Natural Language Processing and the 9th International Joint Conference on
  Natural Language Processing (EMNLP-IJCNLP)}, pages 2153--2162, Hong Kong,
  China. Association for Computational Linguistics.

\bibitem[{Wallace et~al.(2019{\natexlab{b}})Wallace, Rodriguez, Feng, Yamada,
  and Boyd-Graber}]{wallace-etal-2019-trick}
Eric Wallace, Pedro Rodriguez, Shi Feng, Ikuya Yamada, and Jordan Boyd-Graber.
  2019{\natexlab{b}}.
\newblock \href {https://doi.org/10.1162/tacl_a_00279} {Trick me if you can:
  Human-in-the-loop generation of adversarial examples for question answering}.
\newblock \emph{Transactions of the Association for Computational Linguistics},
  7:387--401.

\bibitem[{Wang et~al.(2019{\natexlab{a}})Wang, Pruksachatkun, Nangia, Singh,
  Michael, Hill, Levy, and Bowman}]{wang2019superglue}
Alex Wang, Yada Pruksachatkun, Nikita Nangia, Amanpreet Singh, Julian Michael,
  Felix Hill, Omer Levy, and Samuel Bowman. 2019{\natexlab{a}}.
\newblock Superglue: A stickier benchmark for general-purpose language
  understanding systems.
\newblock \emph{Advances in neural information processing systems}, 32.

\bibitem[{Wang et~al.(2018)Wang, Singh, Michael, Hill, Levy, and
  Bowman}]{wang-etal-2018-glue}
Alex Wang, Amanpreet Singh, Julian Michael, Felix Hill, Omer Levy, and Samuel
  Bowman. 2018.
\newblock \href {https://doi.org/10.18653/v1/W18-5446} {{GLUE}: A multi-task
  benchmark and analysis platform for natural language understanding}.
\newblock In \emph{Proceedings of the 2018 {EMNLP} Workshop {B}lackbox{NLP}:
  Analyzing and Interpreting Neural Networks for {NLP}}, pages 353--355,
  Brussels, Belgium. Association for Computational Linguistics.

\bibitem[{Wang and Komatsuzaki(2021)}]{gpt-j}
Ben Wang and Aran Komatsuzaki. 2021.
\newblock {GPT-J-6B: A 6 Billion Parameter Autoregressive Language Model}.
\newblock \url{https://github.com/kingoflolz/mesh-transformer-jax}.

\bibitem[{Wang et~al.(2019{\natexlab{b}})Wang, Wang, Wang, Wang, and
  Ye}]{wang2019towards}
Wenqi Wang, Run Wang, Lina Wang, Zhibo Wang, and Aoshuang Ye.
  2019{\natexlab{b}}.
\newblock Towards a robust deep neural network in texts: A survey.
\newblock \emph{arXiv preprint arXiv:1902.07285}.

\bibitem[{Wei et~al.(2022)Wei, Tay, Bommasani, Raffel, Zoph, Borgeaud,
  Yogatama, Bosma, Zhou, Metzler et~al.}]{wei2022emergent}
Jason Wei, Yi~Tay, Rishi Bommasani, Colin Raffel, Barret Zoph, Sebastian
  Borgeaud, Dani Yogatama, Maarten Bosma, Denny Zhou, Donald Metzler, et~al.
  2022.
\newblock Emergent abilities of large language models.
\newblock \emph{arXiv preprint arXiv:2206.07682}.

\bibitem[{Wolf et~al.(2020)Wolf, Debut, Sanh, Chaumond, Delangue, Moi, Cistac,
  Rault, Louf, Funtowicz, Davison, Shleifer, von Platen, Ma, Jernite, Plu, Xu,
  Le~Scao, Gugger, Drame, Lhoest, and Rush}]{wolf-etal-2020-transformers}
Thomas Wolf, Lysandre Debut, Victor Sanh, Julien Chaumond, Clement Delangue,
  Anthony Moi, Pierric Cistac, Tim Rault, Remi Louf, Morgan Funtowicz, Joe
  Davison, Sam Shleifer, Patrick von Platen, Clara Ma, Yacine Jernite, Julien
  Plu, Canwen Xu, Teven Le~Scao, Sylvain Gugger, Mariama Drame, Quentin Lhoest,
  and Alexander Rush. 2020.
\newblock \href {https://doi.org/10.18653/v1/2020.emnlp-demos.6} {Transformers:
  State-of-the-art natural language processing}.
\newblock In \emph{Proceedings of the 2020 Conference on Empirical Methods in
  Natural Language Processing: System Demonstrations}, pages 38--45, Online.
  Association for Computational Linguistics.

\bibitem[{Wortsman et~al.(2021)Wortsman, Ilharco, Li, Kim, Hajishirzi, Farhadi,
  Namkoong, and Schmidt}]{wortsman2021robust}
Mitchell Wortsman, Gabriel Ilharco, Mike Li, Jong~Wook Kim, Hannaneh
  Hajishirzi, Ali Farhadi, Hongseok Namkoong, and Ludwig Schmidt. 2021.
\newblock Robust fine-tuning of zero-shot models.
\newblock \emph{arXiv preprint arXiv:2109.01903}.

\bibitem[{Wu et~al.(2021)Wu, Arendt, and Volkova}]{wu-etal-2021-evaluating}
Winston Wu, Dustin Arendt, and Svitlana Volkova. 2021.
\newblock \href {https://doi.org/10.18653/v1/2021.eacl-main.210} {Evaluating
  neural model robustness for machine comprehension}.
\newblock In \emph{Proceedings of the 16th Conference of the European Chapter
  of the Association for Computational Linguistics: Main Volume}, pages
  2470--2481, Online. Association for Computational Linguistics.

\bibitem[{Yang et~al.(2018)Yang, Qi, Zhang, Bengio, Cohen, Salakhutdinov, and
  Manning}]{yang-etal-2018-hotpotqa}
Zhilin Yang, Peng Qi, Saizheng Zhang, Yoshua Bengio, William Cohen, Ruslan
  Salakhutdinov, and Christopher~D. Manning. 2018.
\newblock \href {https://doi.org/10.18653/v1/D18-1259} {{H}otpot{QA}: A dataset
  for diverse, explainable multi-hop question answering}.
\newblock In \emph{Proceedings of the 2018 Conference on Empirical Methods in
  Natural Language Processing}, pages 2369--2380, Brussels, Belgium.
  Association for Computational Linguistics.

\bibitem[{Yogatama et~al.(2019)Yogatama, d'Autume, Connor, Kocisky,
  Chrzanowski, Kong, Lazaridou, Ling, Yu, Dyer et~al.}]{yogatama2019learning}
Dani Yogatama, Cyprien de~Masson d'Autume, Jerome Connor, Tomas Kocisky, Mike
  Chrzanowski, Lingpeng Kong, Angeliki Lazaridou, Wang Ling, Lei Yu, Chris
  Dyer, et~al. 2019.
\newblock Learning and evaluating general linguistic intelligence.
\newblock \emph{arXiv preprint arXiv:1901.11373}.

\bibitem[{Zhang et~al.(2022)Zhang, Roller, Goyal, Artetxe, Chen, Chen, Dewan,
  Diab, Li, Lin et~al.}]{zhang2022opt}
Susan Zhang, Stephen Roller, Naman Goyal, Mikel Artetxe, Moya Chen, Shuohui
  Chen, Christopher Dewan, Mona Diab, Xian Li, Xi~Victoria Lin, et~al. 2022.
\newblock Opt: Open pre-trained transformer language models.
\newblock \emph{arXiv preprint arXiv:2205.01068}.

\bibitem[{Zhu et~al.(2020)Zhu, Cheng, Gan, Sun, Goldstein, and
  Liu}]{Zhu2020FreeLB}
Chen Zhu, Yu~Cheng, Zhe Gan, Siqi Sun, Tom Goldstein, and Jingjing Liu. 2020.
\newblock \href {https://openreview.net/forum?id=BygzbyHFvB} {Freelb: Enhanced
  adversarial training for natural language understanding}.
\newblock In \emph{International Conference on Learning Representations}.

\end{thebibliography}
\clearpage
\appendix

\section{Appendix}
\label{sec:appendix}
\begin{figure}
    \includegraphics[width=\textwidth]{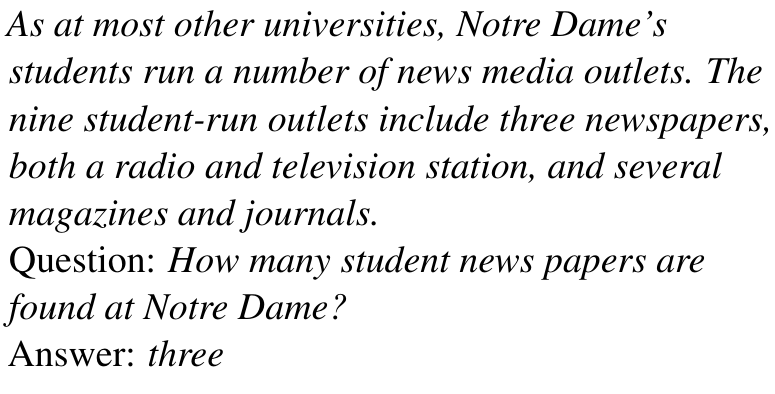}
    \caption{A sample from SQuAD with the input formatting used for fine-tuning decoder-only models, in-context learning, and zero-shot inference.}
    \label{fig:prompt}
\end{figure}

\subsection{Training Details}
\label{traning details}
In addition to sharing hyperparameters for each model, we plan to share all model weights on the HuggingFace Hub~\cite{wolf-etal-2020-transformers} such that the community can continue to evaluate the models in our testbed. 
\subsubsection{Span prediction fine-tuning}
\label{encoder only fully finetuned params}
We fine-tune models by adding a span prediction head and fine-tuning for 2 epochs using a learning rate of 3e-5 and a linear learning rate decay.
\subsubsection{Prompt fine-tuning}
\textbf{Encoder-Decoder Models}\newline
We fine-tune encoder-decoder models on both question->answer generation (mask filling) and answer generation tasks from ~\citet{chada-natarajan-2021-fewshotqa}. For the question->answer generation task we fine-tune the models for 2 epochs, use a linear learning rate decay, and search for the best learning rate from 1e-4, 5e-5, and 3e-5 based on performance on the validation set. For the answer generation task we fine-tune for 2 epochs with a learning rate of 3e-5 and linear learning rate decay.\newline
\textbf{Decoder-Only Models}\newline
We fine-tune decoder only models using a language modeling head. Specifically we format samples as shown in Figure~\ref{fig:prompt} and only calculate loss on the answer tokens. We search for the best learning rate among 5e-5 and 5e-6 and use a linear learning rate decay. In addition we search for the best weight decay value between 0.01 and 0.1. We fine-tune these models for 5 epochs and pick the model with the best validation set F1 score.

\subsubsection{Parameter efficient fine-tuning}
\textbf{Fine-tuned Models}\newline
As suggested in the Adapter-Transformers library~\cite{pfeiffer-etal-2020-adapterhub} we use a learning rate of 1e-4 with linear learning rate decay and fine-tune for 15 epochs for all parameter efficient fine-tuning methods picking the model with the best F1 score on the validation set.\newline
\textbf{Prompt Fine-tuned Models}\newline
We fine-tune using a learning rate of 1e-4 with linear learning rate decay and 10 epochs for all parameter efficient fine-tuning methods picking the model with the best F1 score on the validation set.

\subsubsection{Few-shot Fine-tuning}
We fine-tune models on $2^4$ to $2^{10}$ samples from SQuAD (doubling the size as we increase the number of shots). We repeat each experiment three times using randomly picked samples to remove outliers that result from fine-tuning on specific examples.\newline
\textbf{Fine-tuned Models}\newline
We use the same fine-tuning setup as \citet{ram-etal-2021-shot}. Specifically we fine-tune for 10 epochs or 200 steps (picking which ever is largest). We use a learning rate of 3e-5 with a linear learning rate decay and 0.1 warm-up ratio.\newline
\textbf{Prompt Fine-tuned Models}\newline
For autoregressive models we fine-tune for 10 epochs with a learning rate of 1e-5 and linear learning rate decay. In addition we use a weight decay of 0.1 as we find in our experiments fine-tuning decoder-only models that this is an ideal value for the models we evaluate. For T5 and BART we use the same evaluation setup as \citet{chada-natarajan-2021-fewshotqa} for both masked span prediction and answer generation methods.
\subsubsection{Robustness enhanced methods}
\textbf{RXF}\newline
We adapt the official implementation\footnote{https://github.com/facebookresearch/fairseq/tree/main/examples/rxf} for RXF (using the R3F variant) to fine-tune encoder only question answering models. We use the same fine-tuning hyper-parameters as the fully fine-tuned encoder only models~\ref{encoder only fully finetuned params} but use polynomial learning rate decay, weight decay value of 0.01, and warm-up ratio of 0.06. For R3F specific parameters we use $\lambda$=1.0, $\delta$=1e-5, and Normal noise type.\newline
\textbf{FreeLB}\newline
We use the official implementation\footnote{https://github.com/zhuchen03/FreeLB} for FreeLB to fine-tune encoder only question answering models. We use a learning rate of 5e-6 and fine-tune for 2 epochs with linear learning rate decay. For FreeLB specific parameters we set $m = 2$, $\alpha$=1e-1, and $\epsilon$=6e-1.

\subsubsection{Zero-shot inference}
\label{sec:zeroshot setup}
For zero-shot evaluations we pre-process each sample into the format in Figure~\ref{fig:prompt} omitting the answer from the prompt. We generate the answer using beam decoding with five beams for models smaller than 2 billion parameters and use greedy decoding for the rest of the models. We use a maximum generation length of 20 tokens and use the end of sequence token to terminate generation.

\subsubsection{In-context learning}
For in-context learning evaluations we condition a language model on one or four random samples examples from the SQuAD training set. Figure~\ref{fig:prompt} illustrates the format of each sample. When we are conditioning on multiple samples we separate each formatted sample with a newline delimiter. For each training sample we condition on, we truncate the context of the sample to 100 tokens. Furthermore, we truncate the context of input sample (sample we are running inference on) to 200 tokens. For each model and number of shots we repeat each experiment three times using randomly samples training shots. The exception to this is the OPT 175 billion parameter model which we evaluate only once in the one and four shot settings. We use the same generation setup as zero-shot inference~\ref{sec:zeroshot setup}.

\subsection{Additional Plots}
\begin{figure}
    \centering
    \includegraphics[width=0.9\columnwidth]{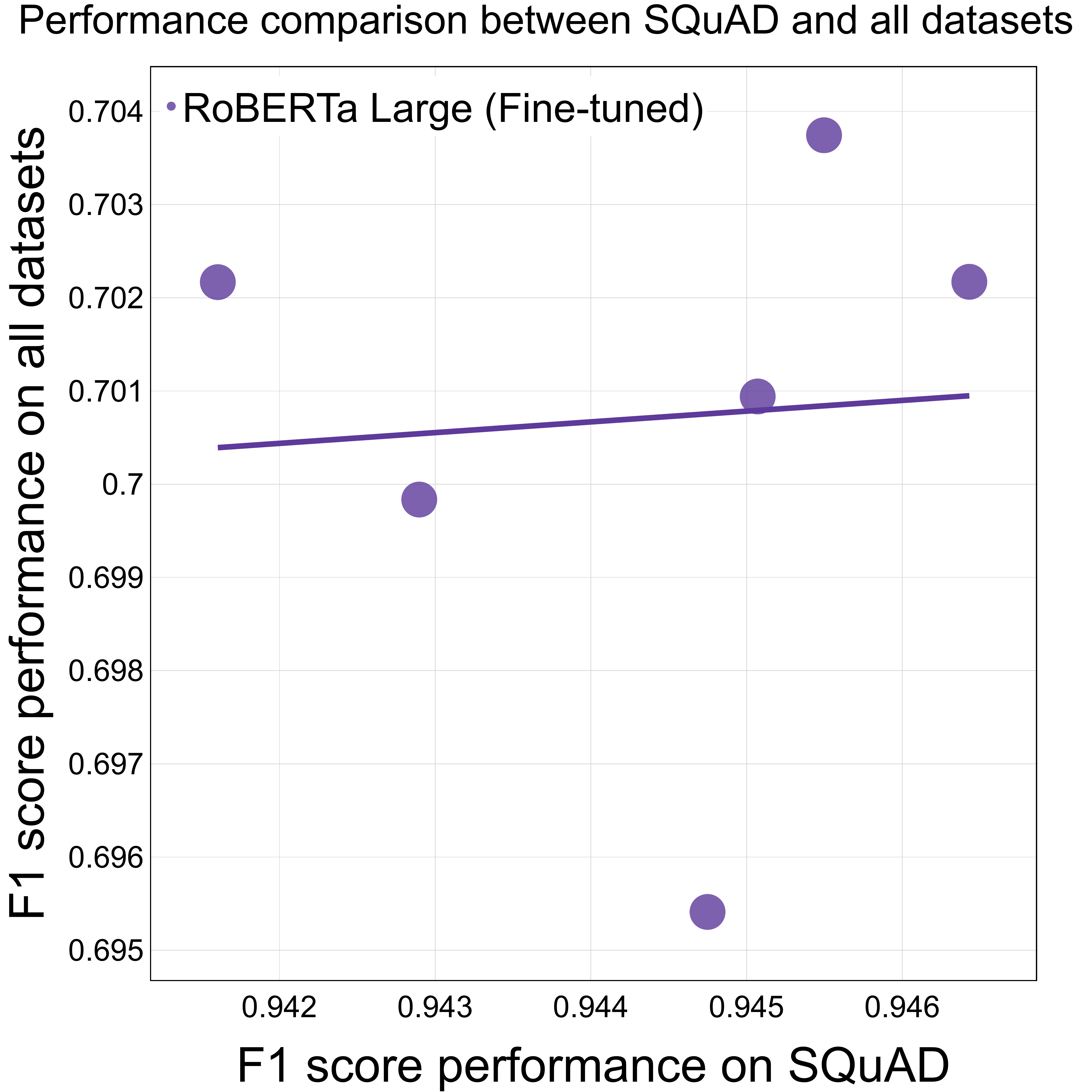}
    \caption{The average effective robustness for six fine-tuning runs of RoBERTa Large shows that the robustness differences  between fine-tuning runs are negligible.}
    \label{fig:runs}
\end{figure}
In this section we include Figure~\ref{fig:runs} which shows the effect of different fine-tuning runs on effective robustness. We find that even when we fine-tune six different RoBERTa Large models by varying data ordering and weight initialization for the span prediction head the average effective robustness on all distribution shifts is stable.

\newpage

\begin{figure*}
     \centering
     \begin{subfigure}[b]{0.25\textwidth}
         \centering
         \includegraphics[width=\textwidth]{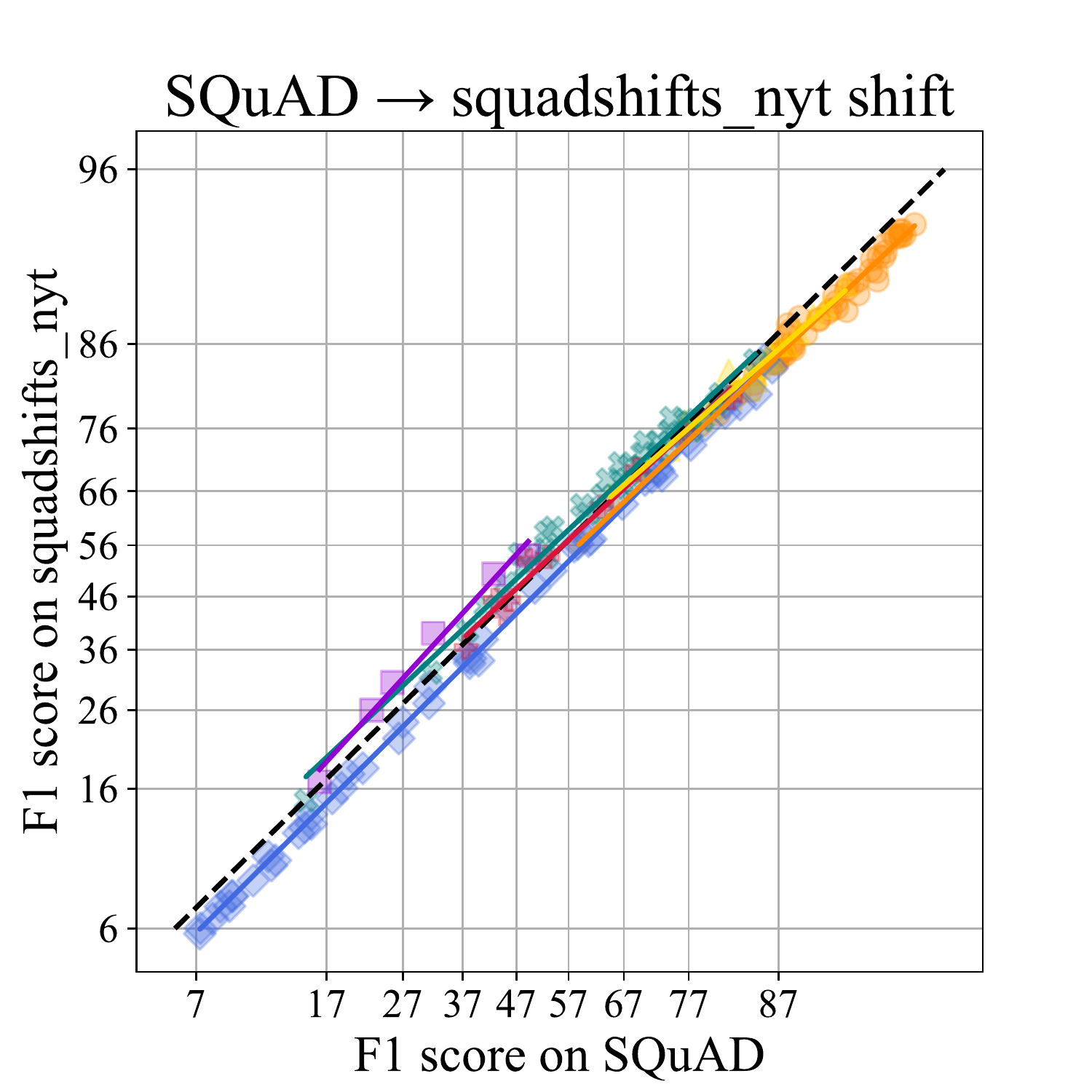}
         \caption{SquadShifts NYT}
         \label{fig:all shifts nyt}
     \end{subfigure}
     \begin{subfigure}[b]{0.25\textwidth}
         \centering
         \includegraphics[width=\textwidth]{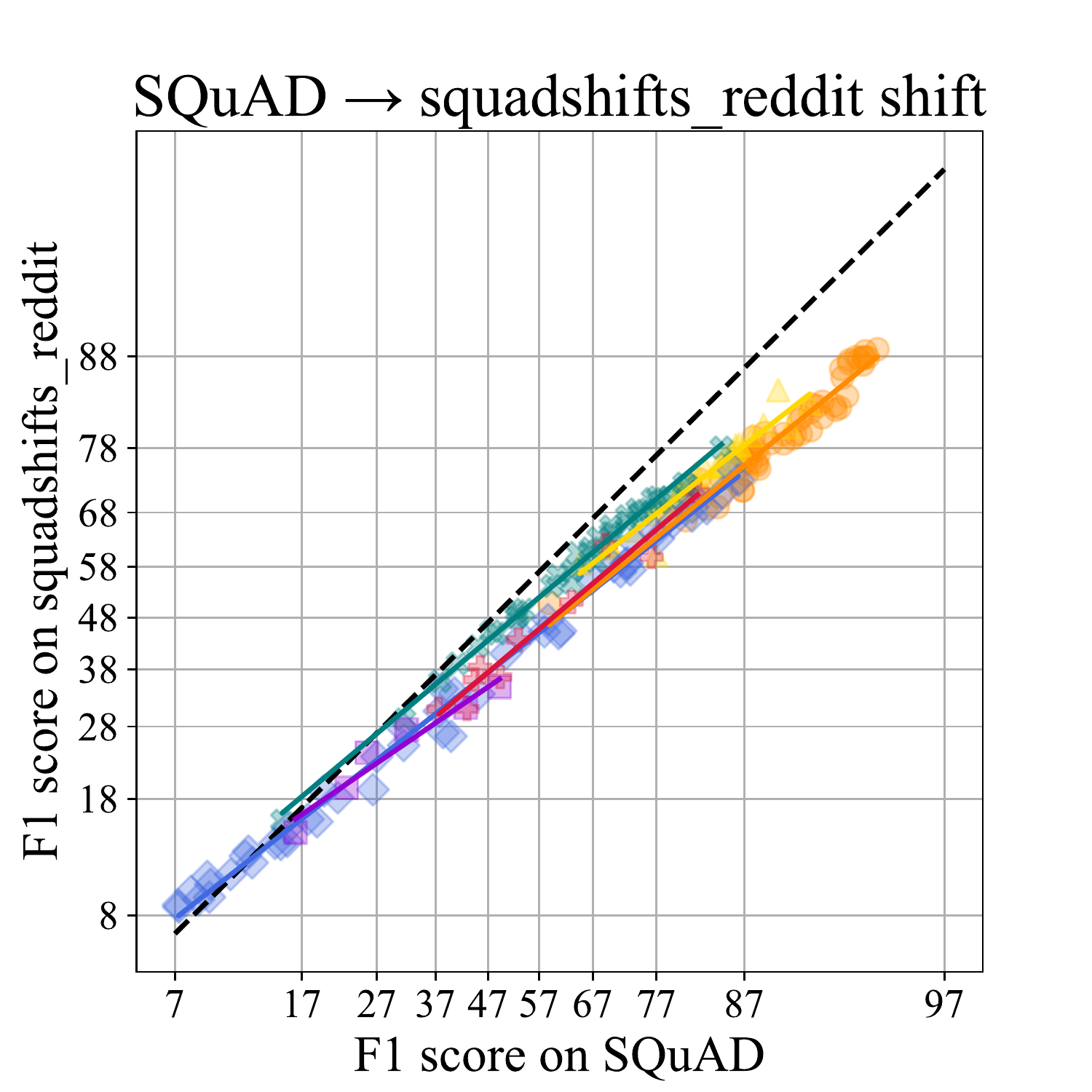}
         \caption{SquadShifts Reddit}
         \label{fig:all shifts reddit}
     \end{subfigure}
     \begin{subfigure}[b]{0.25\textwidth}
         \centering
         \includegraphics[width=\textwidth]{figures/SQuAD_to_squadshifts_new_wiki_mpl.pdf}
         \caption{SquadShifts Wikipedia}
     \end{subfigure}
     \begin{subfigure}[b]{0.25\textwidth}
         \centering
         \includegraphics[width=\textwidth]{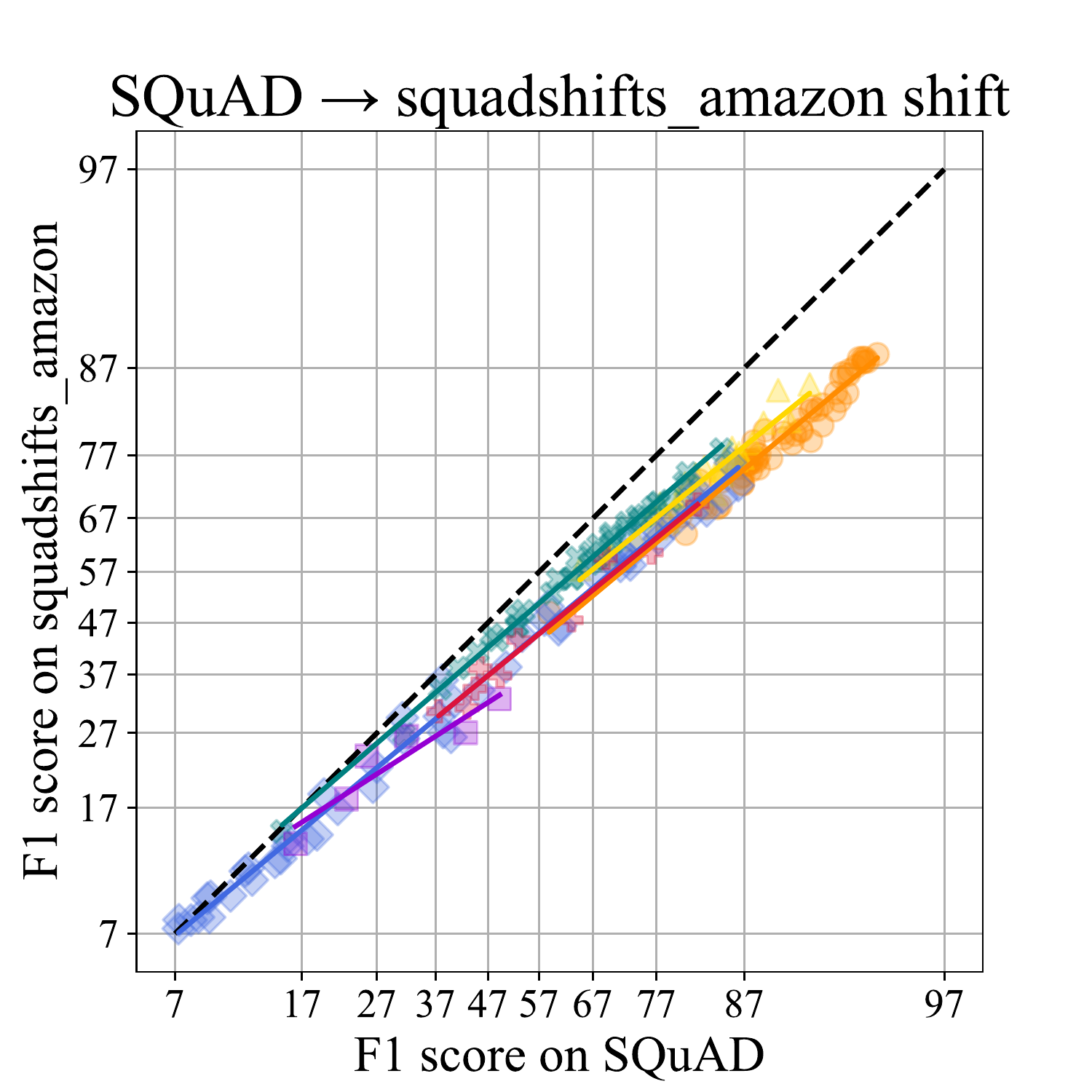}
         \caption{SquadShifts Amazon}
         \label{fig:all shifts amazon}
     \end{subfigure}
     \begin{subfigure}[b]{0.25\textwidth}
         \centering
         \includegraphics[width=\textwidth]{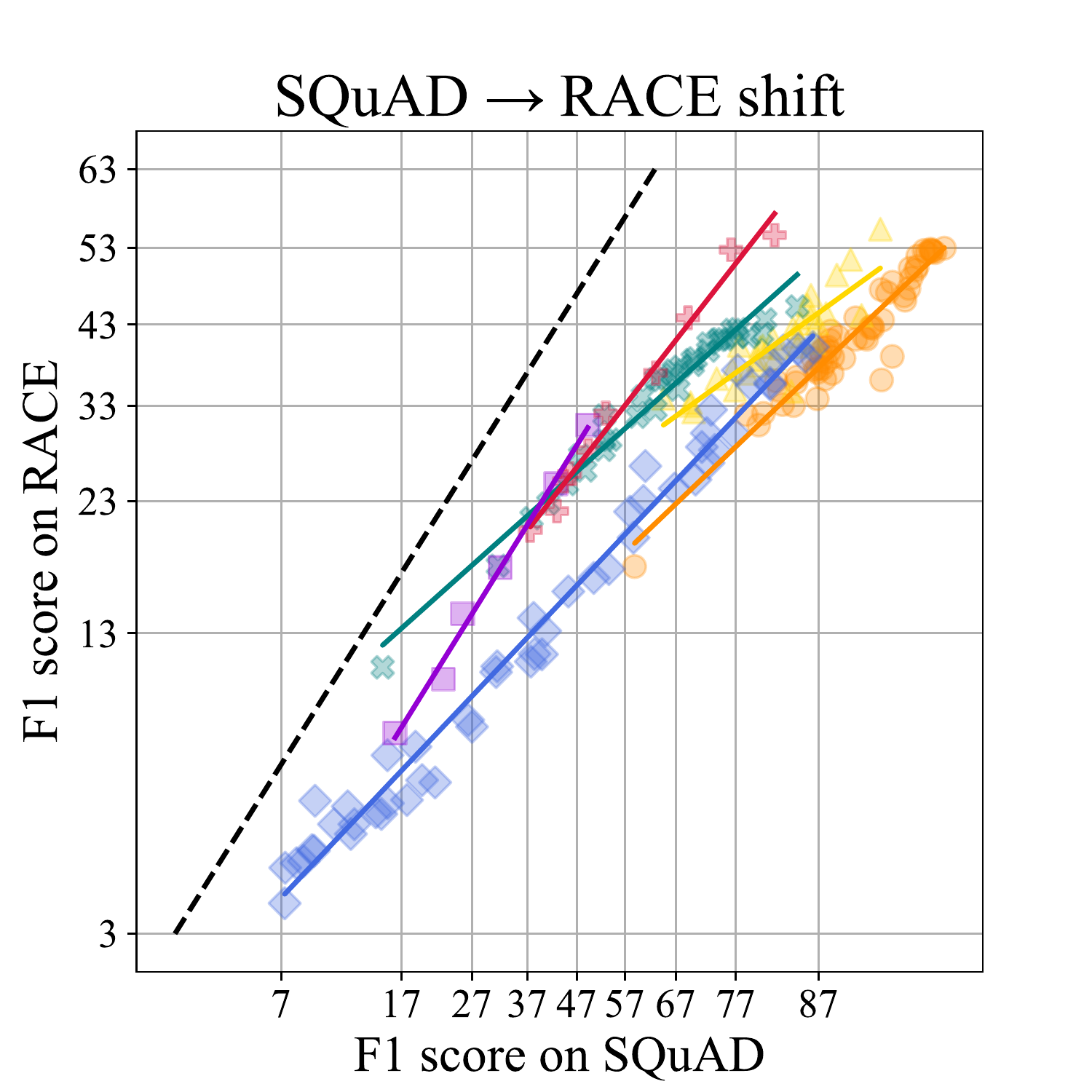}
         \caption{RACE}
         \label{fig:all shifts race}
     \end{subfigure}
     \begin{subfigure}[b]{0.25\textwidth}
         \centering
         \includegraphics[width=\textwidth]{figures/SQuAD_to_DROP_mpl.pdf}
         \caption{DROP}
     \end{subfigure}
     \begin{subfigure}[b]{0.25\textwidth}
         \centering
         \includegraphics[width=\textwidth]{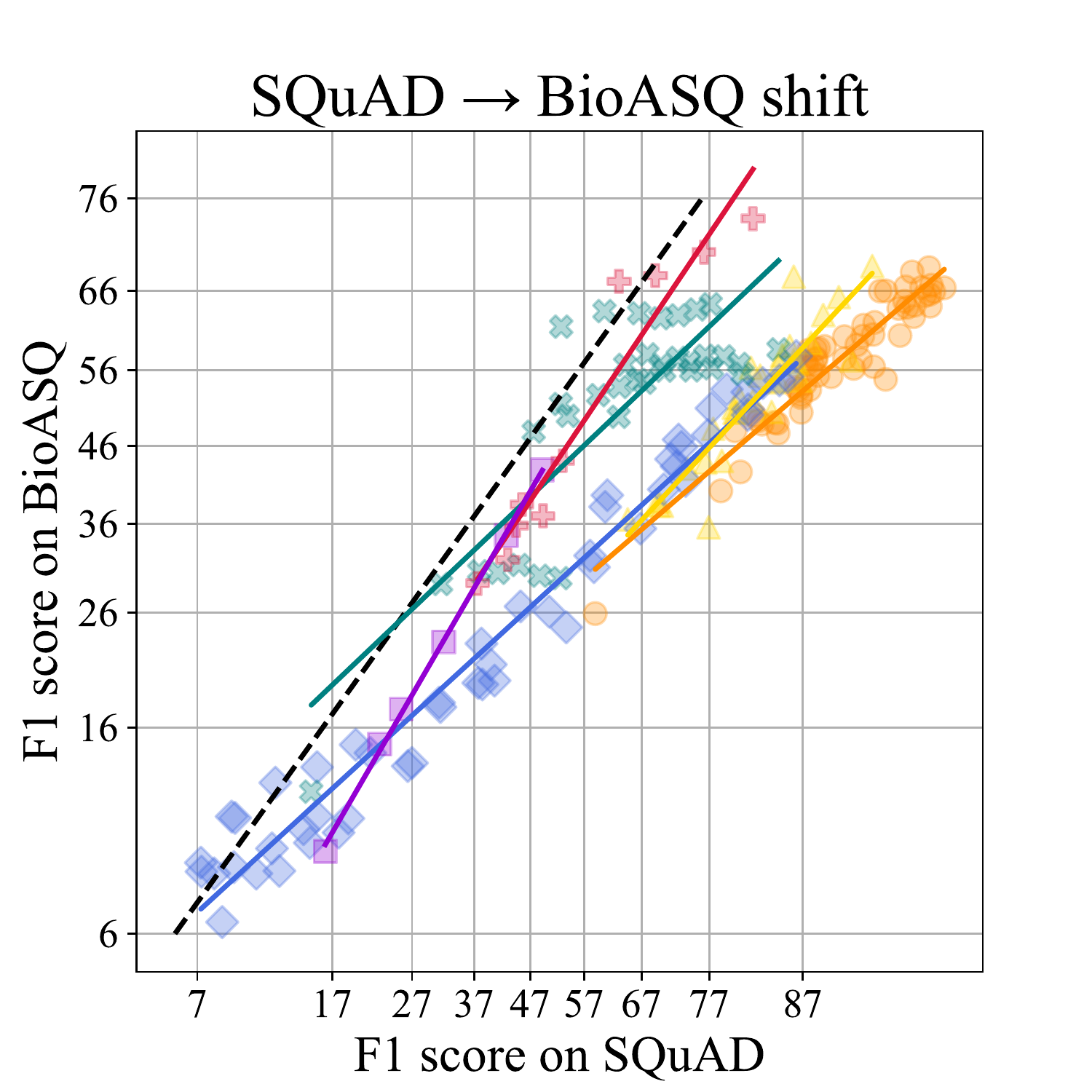}
         \caption{BioASQ}
         \label{fig:all shifts bioasq}
     \end{subfigure}
     \begin{subfigure}[b]{0.25\textwidth}
         \centering
         \includegraphics[width=\textwidth]{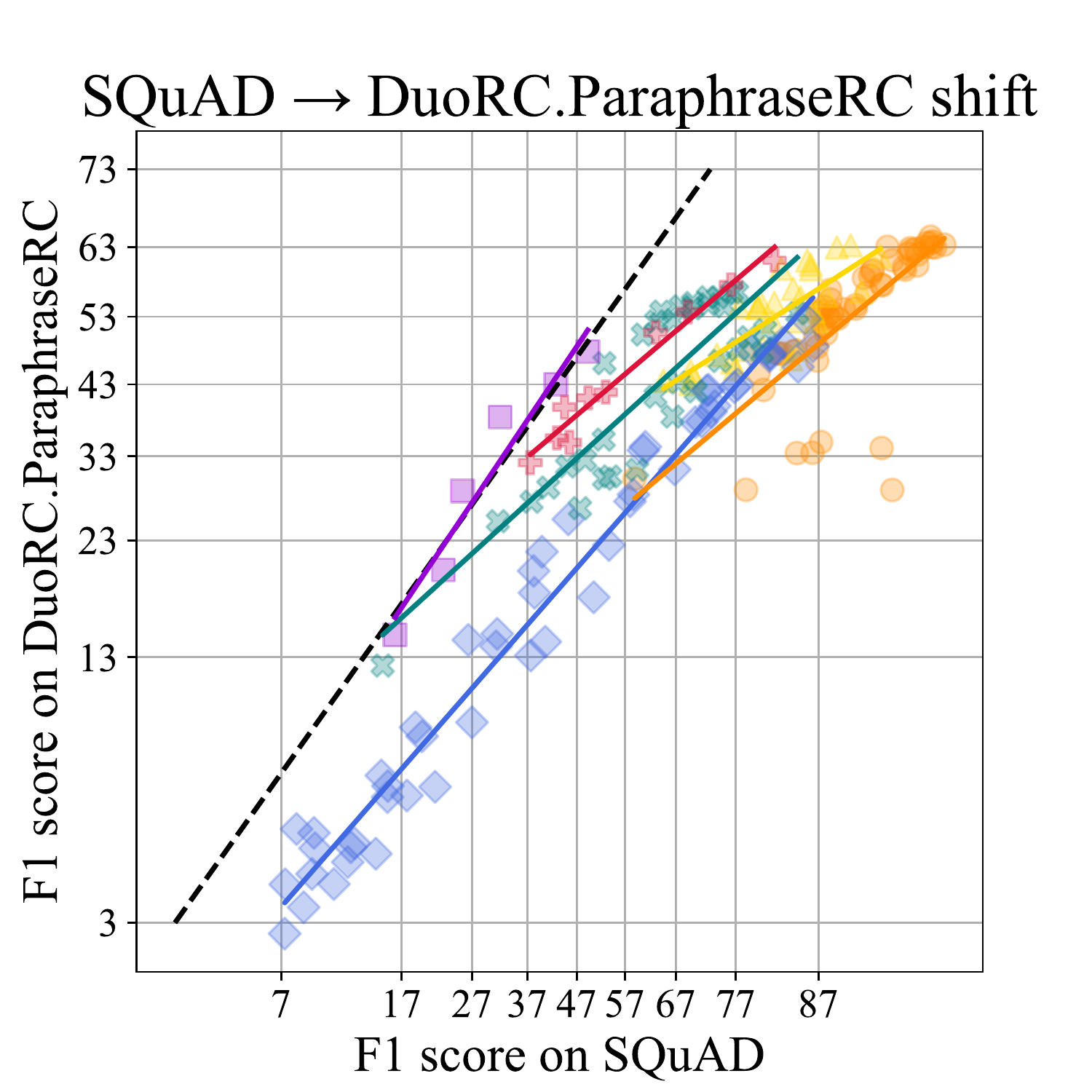}
         \caption{DuoRC}
         \label{fig:all shifts duorc}
     \end{subfigure}
     \begin{subfigure}[b]{0.25\textwidth}
         \centering
         \includegraphics[width=\textwidth]{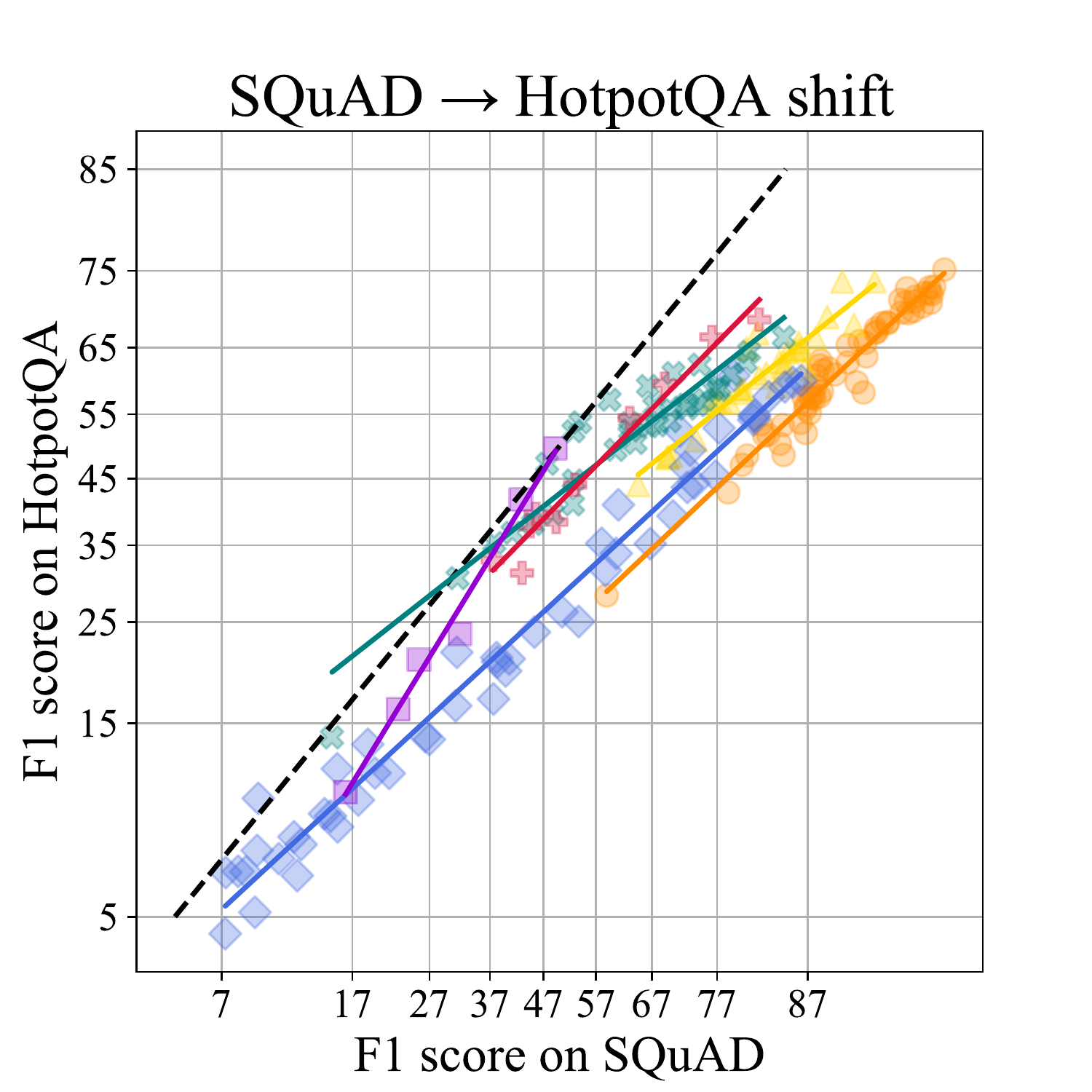}
         \caption{HotpotQA}
         \label{fig:all shifts hotpotqa}
     \end{subfigure}
     \begin{subfigure}[b]{0.25\textwidth}
         \centering
         \includegraphics[width=\textwidth]{figures/SQuAD_to_SearchQA_mpl.pdf}
         \caption{SearchQA}
     \end{subfigure}
     \begin{subfigure}[b]{0.25\textwidth}
         \centering
         \includegraphics[width=\textwidth]{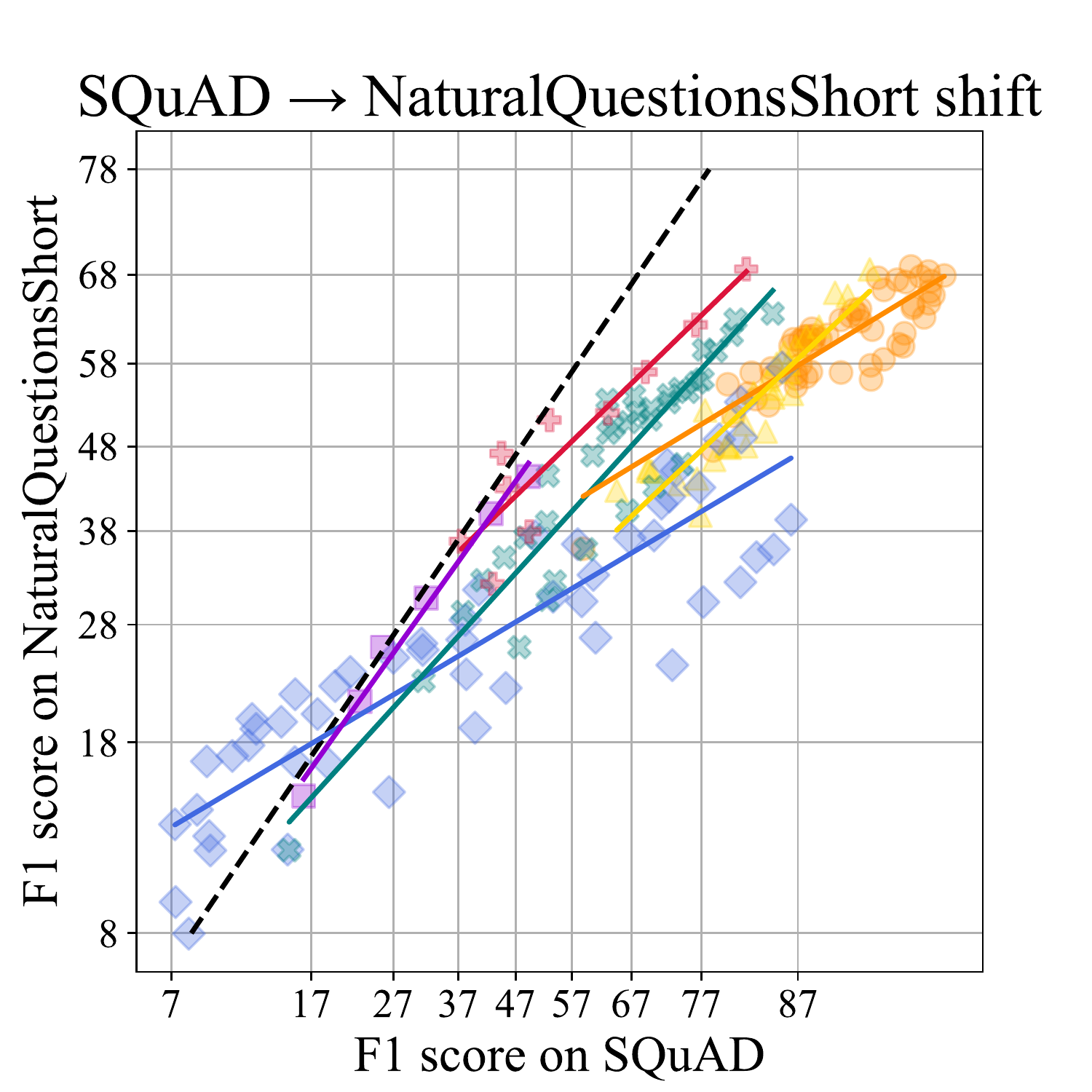}
         \caption{Natural Questions}
         \label{fig:all shifts natural questions}
     \end{subfigure}
     \begin{subfigure}[b]{0.25\textwidth}
         \centering
         \includegraphics[width=\textwidth]{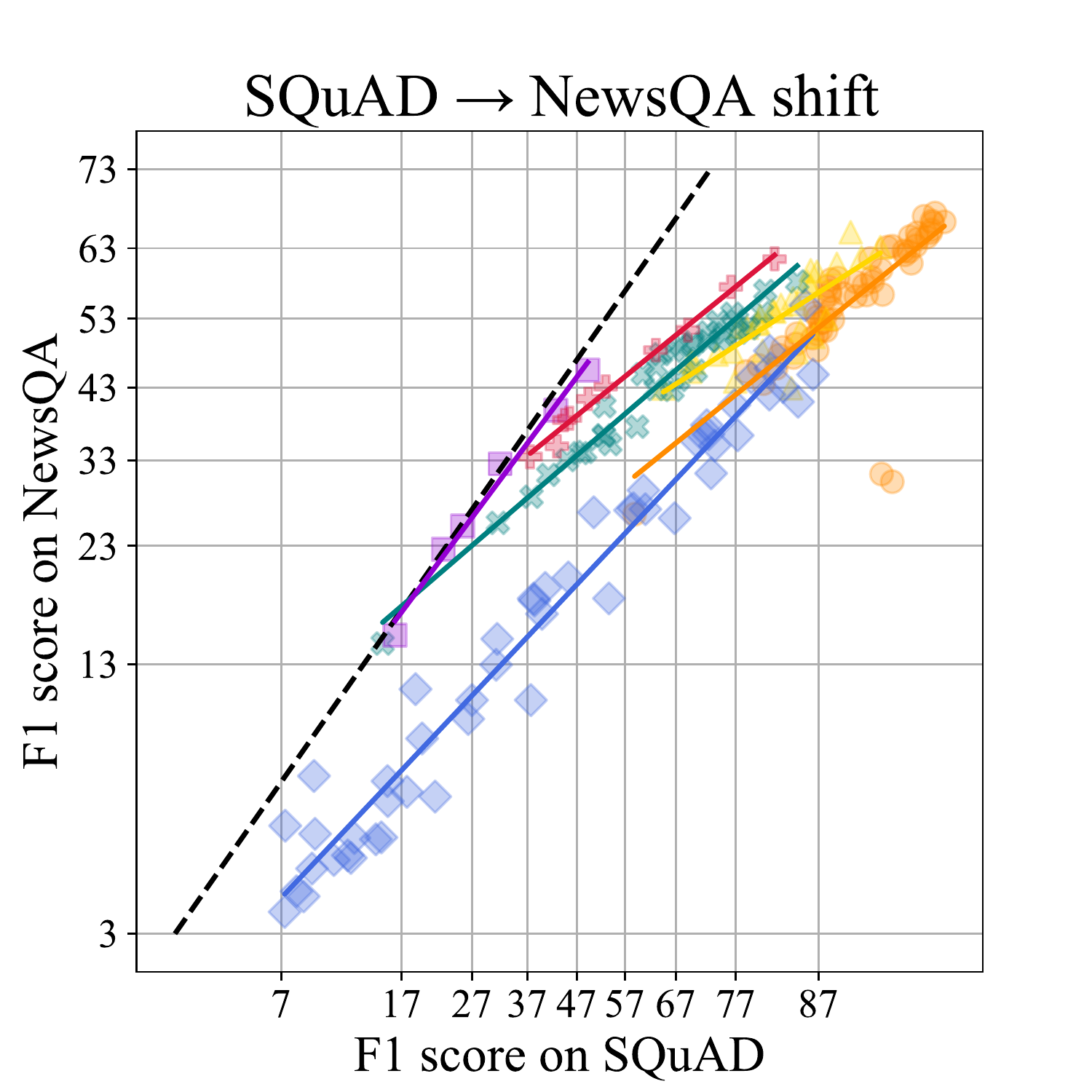}
         \caption{NewsQA}
         \label{fig:all shifts newsqa}
     \end{subfigure}
     \begin{subfigure}[b]{0.25\textwidth}
         \centering
         \includegraphics[width=\textwidth]{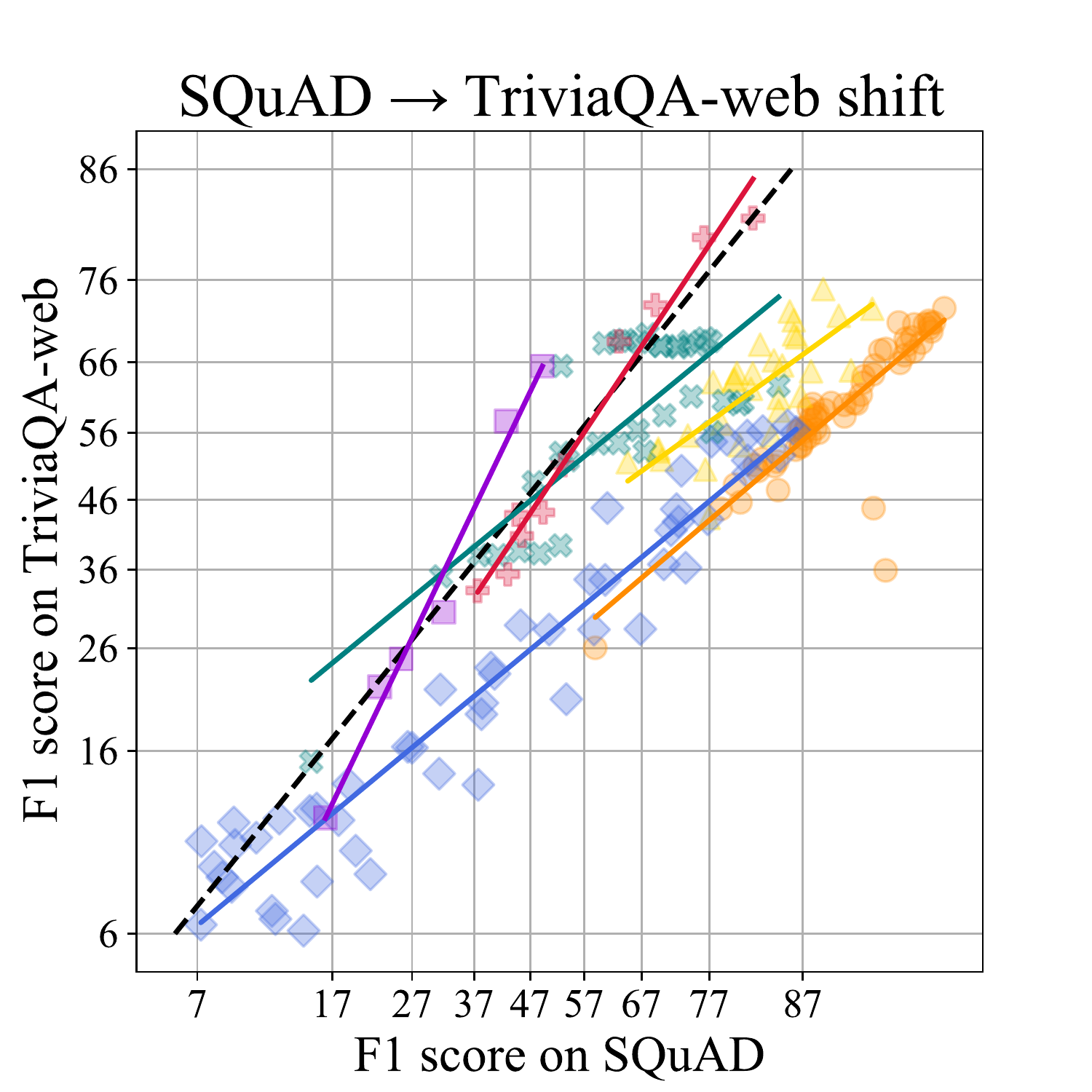}
         \caption{TriviaQA}
         \label{fig:all shifts triviaqa}
     \end{subfigure}
     \begin{subfigure}[b]{0.25\textwidth}
         \centering
         \includegraphics[width=\textwidth]{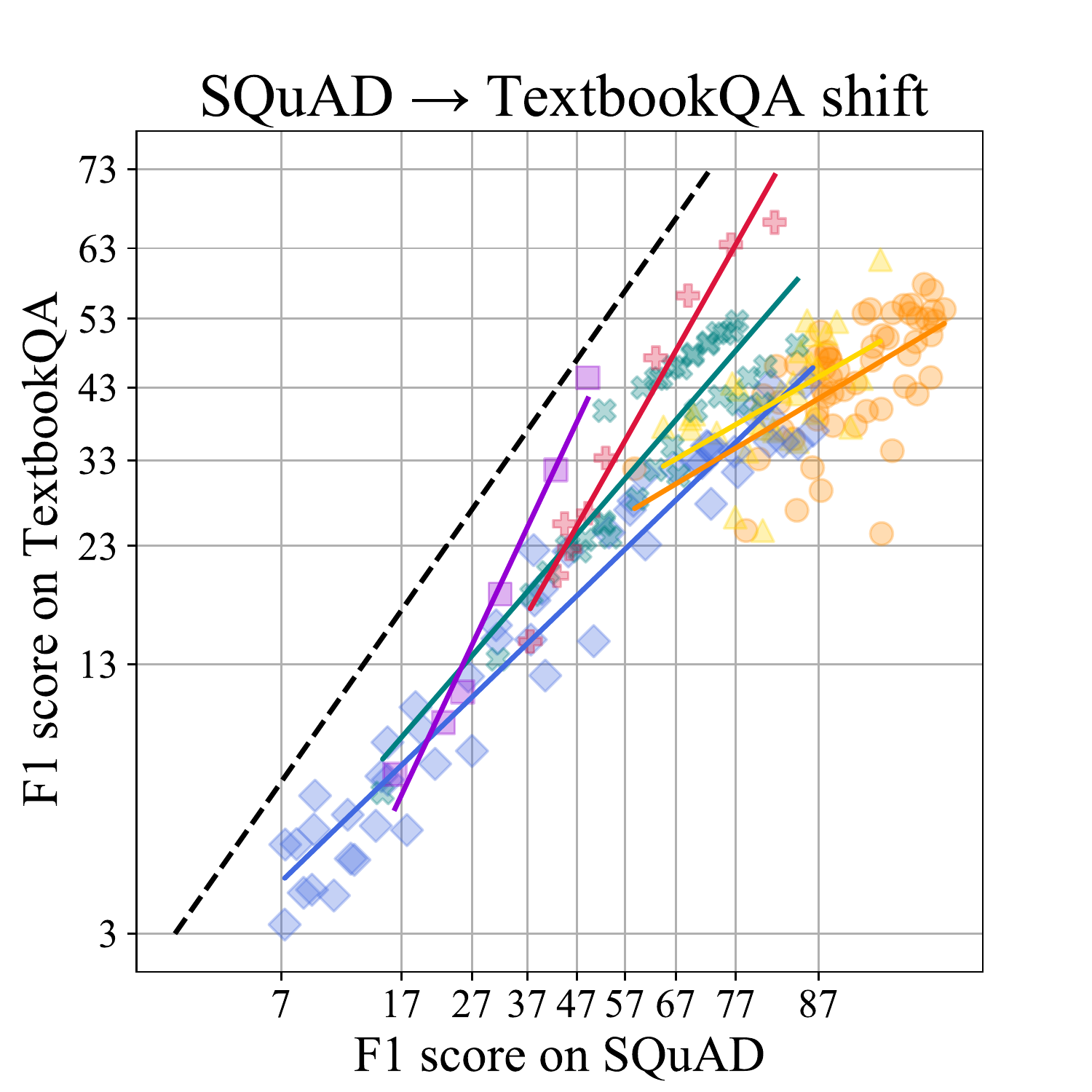}
         \caption{TextbookQA}
         \label{fig:all shifts textbookqa}
     \end{subfigure}
     \begin{subfigure}[b]{0.25\textwidth}
         \centering
         \includegraphics[width=\textwidth]{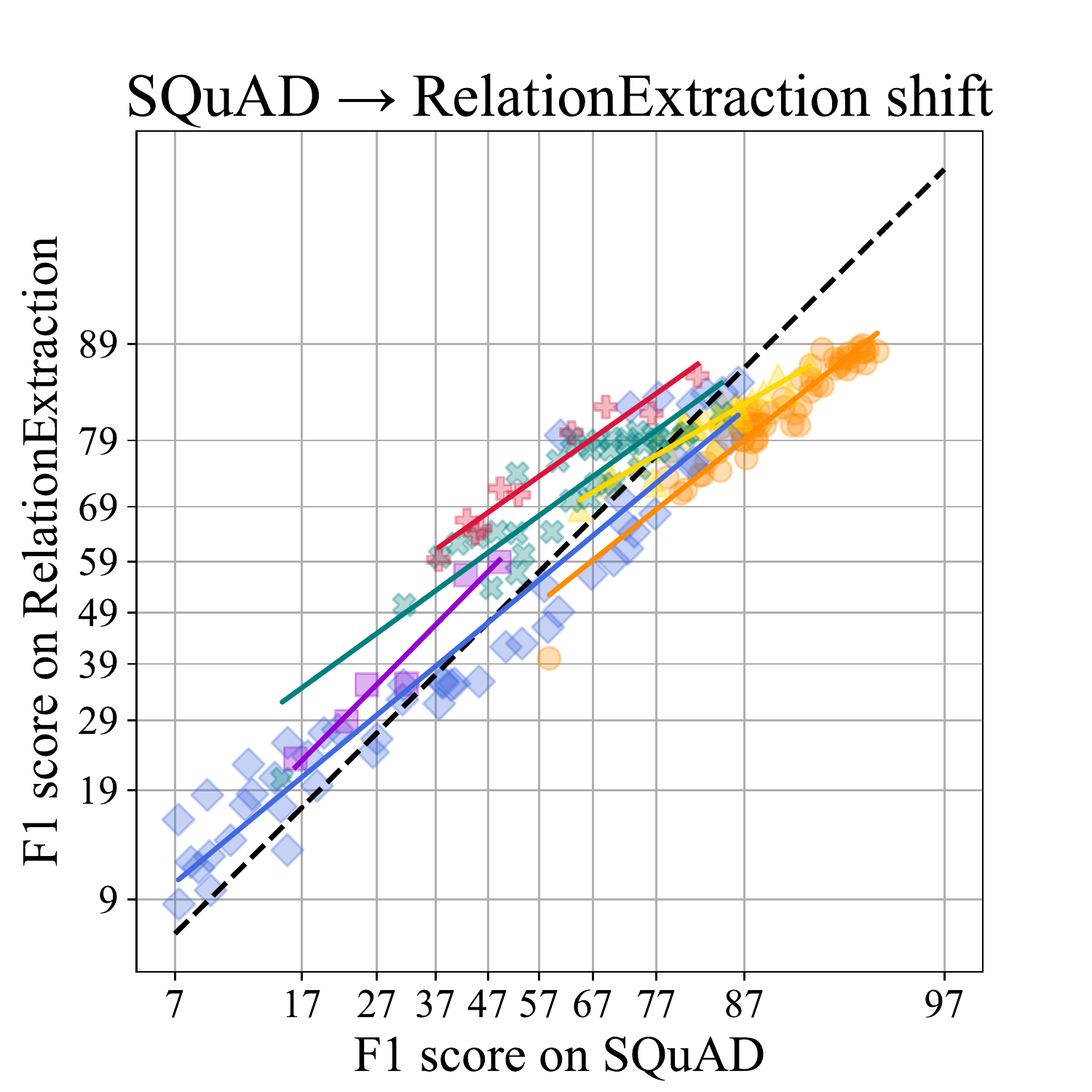}
         \caption{Relation Extraction}
        \label{fig:all shifts relation extraction}
     \end{subfigure}
     \begin{subfigure}[b]{0.4\textwidth}
         \centering
         \vspace{0.5cm}
         \includegraphics[width=\textwidth]{figures/legend-mpl.pdf}
     \end{subfigure}
     \caption{Instead of averaging over all 15 datasets, we show logit-scaled plots examining all 15 distribution shifts individually.}
     \label{fig:all shifts}
\end{figure*}

\end{document}